\documentclass[11pt]{article}

\usepackage[final]{acl}

\usepackage{times}
\usepackage{latexsym}
\usepackage{tcolorbox}
\usepackage{enumitem} 

\usepackage{amsfonts}

\usepackage[T1]{fontenc}

\usepackage[utf8]{inputenc}
\usepackage{amsmath}

\usepackage{microtype}

\usepackage{inconsolata}
\usepackage{graphicx}
\usepackage{lipsum}

\usepackage{graphicx}
\usepackage{booktabs}
\usepackage[table]{xcolor} 
\newcommand{\header}[1]{\textbf{#1}}
\newcommand{\oursrow}{\rowcolor{gray!12}}

\title{A Universal Avoidance Method for Diverse Multi-branch Generation}

\author{Kyeongman Park \\
  Seoul National University \\
  \texttt{zzangmane@snu.ac.kr} \\\And
  Minha Jhang \\
  Seoul National University \\
  \texttt{jminha@snu.ac.kr} \\\And
  Kyomin Jung \\
  Seoul National University \\
  \texttt{kjung@snu.ac.kr}}

\begin{document}
\maketitle
\begin{abstract}
Modern generative models still lack human-level creativity, particularly in multi-branch diversity. Prior approaches to address this problem often incur heavy computation or strong dependency on model architecture. Therefore, we introduce \textbf{UAG}(\textbf{U}niversal \textbf{A}voidance \textbf{G}eneration), a model-agnostic and computationally efficient generation strategy that penalizes similarity among previously generated outputs. Thus, UAG can enhance multi-branch diversity across both diffusion and transformer models, with minimal additional computation. In experiments, our method achieves up to 1.9 times higher diversity, runs 4.4 times faster, and requires only 1/64 of the FLOPs compared to state-of-the-art methods. The full code is  \href{https://anonymous.4open.science/r/2026_ACL_Universal/}{here}.
\end{abstract}

\section{Introduction}

Recently, generative models across various domains such as text and image generation \cite{yang2025qwen3,dubey2024llama,nie2025large,openai2025gpt5system,google_imagen4_2025_blog,openai_dalle3_systemcard_2023,stabilityai_sd3_2024_announce} have demonstrated remarkable performance. Despite this progress, these models still fall short in terms of human-level creativity, particularly in tasks such as story writing \cite{park2024character,wen2023grove,nottingham2024improving,jaschek2019mysterious} and illustration \cite{lu2024procreate,li2024enhancing,zameshina2023diverse}. These tasks require substantial \emph{multi-branch diversity}, i.e., generating diverse outputs from the same prompts \cite{park2025avoidance}.


Thus, many approaches have been proposed to address the lack of \emph{multi-branch diversity} in these tasks \cite{park2025avoidance,nottingham2024improving,wen2023grove,lu2024procreate,li2024enhancing}. However, these approaches often suffer from substantial computational overhead due to token-wise or pixel-wise operations \cite{park2025avoidance,lu2024procreate}, and limited applicability across different models because of their dependence on specific architectures \cite{park2025avoidance,corso2023particle,welleck2019neural,li2024enhancing,sadat2023cads}.
To overcome these limitations, we propose \textbf{UAG}(\textbf{U}niversal \textbf{A}voidance \textbf{G}eneration), a novel framework for enhancing \emph{multi-branch diversity} that minimizes additional computation costs while remaining agnostic to model architecture.

UAG enhances \emph{multi-branch diversity} by reducing similarity to previously generated outputs, by penalizing the gradient of a similarity loss on the final outputs. The similarity loss comprises two components:(1) a local similarity loss at the token or pixel level, and(2) a global similarity loss at the hidden-state level. Through loss scheduling, UAG can promote a natural progression of diversity: in the early stages, local similarity encourages concept-level diversity, while in later stages, global similarity fosters semantic-level diversity. Because UAG requires only a few gradient computations per sample, rather than token-wise \cite{park2025avoidance,welleck2019neural,garces-arias-etal-2024-adaptive,su-collier-2022-contrastive,ding-etal-2025-guard,giulianelli-etal-2023-comes} or pixel-wise operations \cite{ho2020denoising}, it is both highly efficient and fast. Moreover, due to its simplicity, UAG is model-agnostic and thus applicable to both diffusions and transformers, which represent the current state of the art in generative modeling.

Surprisingly, our method achieves on average 1.43 times higher scores in conventional diversity metrics(e.g., BLEU, CLIPscore) and 1.19 times higher scores in LLM-based evaluations compared to baselines. Moreover, our method simultaneously attains 4.4 times faster decoding speed and requires 64 times fewer FLOPs than previous state-of-the-art methods  \cite{park2025avoidance,lu2024procreate}.


\section{Universal Generation Process}
Most modern generative models, including autoregressive and diffusion-based 
models, can be uniformly described as sequential processes indexed 
by steps $t=1,2,\dots,T$. 
At each step $t$, the model updates an internal hidden(or latent) state $h_t$ 
and produces an output representation $y_t$:
\begin{equation}
(y_t, h_{t+1}) \sim P_\theta(\cdot \mid h_t, p),
\end{equation}
where $p$ is the conditioning input(e.g., a text prompt or image condition). We define this process as the Universal Generation Process, and this makes it possible to design 
universal diverse multi-branch generation strategy that operates consistently across different model families.

\paragraph{Autoregressive language models.}
The autoregressive language models include famous LLM such as GPT-5 \cite{openai2025gpt5system}, LLaMa-3B \cite{dubey2024llama}, and Qwen-7B \cite{yang2025qwen3}. For these models, $h_t$ denotes the decoder’s last hidden state after processing tokens up to step $t$, while the model output $y_t$ is the logit vector obtained by applying the output weight matrix to $h_t$. The next state $h_{t+1}$ is updated once a token is predicted and appended.

\paragraph{Diffusion language models.}
For denoising diffusion language models \cite{zhu2025llada,nie2025large}, the process starts from random noise and iteratively refines it into a coherent text representation, where the steps $t$ typically count down from a maximum value $T$ to 1. Here, the state $h_t$ is the noisy latent representation of the text at reverse-time step $t$. The model output $y_t$ is the prediction of the token logits. The next state in the generative sequence, $h_{t-1}$, is then calculated by a scheduler, which uses the current state $h_t$ and the prediction $y_t$ to produce a slightly less noisy representation.

\paragraph{Diffusion image models.}
Similarly, latent diffusion models for image generation \cite{deepmind_imagen3_2024,stabilityai_sd3_2024_announce} also follow a reverse, denoising process within the VAE's compressed latent space. The state $h_t$ is the noisy latent vector at reverse-time step $t$. The model's output $y_t$ is the predicted noise present in $h_t$. The scheduler uses this predicted noise $y_t$ to remove it from the current state $h_t$, thereby computing the next, cleaner state $h_{t-1}$. This process is repeated until the final denoised latent $h_0$ is obtained, which is then decoded by the VAE into the final image.

\section{Universal Avoidance Generation}
To achieve \emph{diverse multi-branch generation} within the Universal Generation Process, we propose \textbf{UAG}(\textbf{U}niversal \textbf{A}voidance \textbf{G}eneration), which applies gradient-based penalties to the output space and adjusts the results step by step. Our method requires only simple differentiation with respect to outputs and hidden states, rather than repetitive token-wise or pixel-wise computation. Thus, it is not only applicable to any generative model within the Universal Generation Process, but also significantly more efficient than previous state-of-the-art methods \cite{park2025avoidance,lu2024procreate}.

\subsection{Step-wise Loss Scheduling}

We denote by $\phi(y_t)$ the output representation of the model.
Let $\mathcal{B}_t^{\text{out}}$ and $\mathcal{B}_t^{\text{hid}}$ be the reference banks, which are previously generated outputs or hidden states from cached past runs.

The step-$t$ local loss  $\mathcal{L}^{(t)}_{\text{local}}$
is defined as $\max_{b\in\mathcal{B}^{\text{out}}_t}
\,\mathrm{sim}\!\big(\phi(y_t),\,b\big)$, and global loss $\mathcal{L}^{(t)}_{\text{global}}$ is defined as $\max_{b\in\mathcal{B}^{\text{hid}}_t}
\,\mathrm{sim}\!\big(h_t,\,b\big)$,
where $\mathrm{sim}(\cdot,\cdot)$ is a similarity measure that depends on the model type(see Appendix \ref{Experimental_Setup}).

To balance the contributions of the two losses, we apply a logistic schedule, following prior work \cite{park2025avoidance} :
\begin{multline}
s_t=\frac{1}{1+e^{\delta\,(t-L_0)}}, \\
w_{\text{local}}(t)=\alpha\,\cdot s_t,\quad
w_{\text{global}}(t)=\beta\,\cdot(1-s_t)
\end{multline}
where $\alpha,\beta$ are maximum weights, $L_0$ is the transition center, and $\delta$ controls sharpness.
Thus, we emphasize \emph{local} similarity(e.g., token-level or pixel-level) in the early stages, while in the later stages we emphasize \emph{global} similarity(e.g., hidden-state level). This scheduling is intuitive, since in the earlier stages the model has not yet formed meaningful global representations and should therefore focus on local contextual diversity, whereas in the later stages it becomes more important to ensure global semantic diversity(e.g., story narratives or scene layouts).

Finally, we define the UAG loss at step $t$ as a combined loss as follows:
\begin{multline}
\mathcal{L}^{(t)}_{\text{UAG}}
= w_{\text{local}}(t)\,\mathcal{L}^{(t)}_{\text{local}}
+ w_{\text{global}}(t)\,\mathcal{L}^{(t)}_{\text{global}}.
\end{multline}

\subsection{Gradient-based Penalty Adaptation}

To penalize similarity to previous outputs, we adjust the output representation $y_t$ by subtracting the gradient of the UAG loss:
\begin{multline}
\hat{y}_{t}
=y_t - \nabla_{y_t}{L}^{(t)}_{\text{UAG}}\\
= y_t - \Big(
w_{\text{local}}(t)\,\nabla_{y_t}\mathcal{L}^{(t)}_{\text{local}}
+ w_{\text{global}}(t)\,\nabla_{y_t}\mathcal{L}^{(t)}_{\text{global}}
\Big)
\end{multline}
then we use $\hat{y}_{t}$ to sample token probabilities or as input for the scheduler of the diffusion model. This penalty introduces a repulsive force that pushes the current generation away from previously generated outputs. For the underlying mathematical motivations of this adjustment, see Appendix \ref{Mathematical Intuition of UAG}.

The local penalty $\nabla_{y_t}\mathcal{L}^{(t)}_{\text{local}}$ and global penalty $\nabla_{y_t}\mathcal{L}^{(t)}_{\text{global}}$ defined as
\begin{equation}
\begin{aligned}
\nabla_{y_t}\mathcal{L}^{(t)}_{\text{local}}
  &= \Big(\tfrac{\partial \phi(y_t)}{\partial y_t}\Big)^{\!\top}
     \nabla_{\phi}\mathcal{L}^{(t)}_{\text{local}}, \\
\nabla_{y_t}\mathcal{L}^{(t)}_{\text{global}}
  &= \Big(\tfrac{\partial h_t}{\partial y_t}\Big)^{\!\top}
     \nabla_{h_{t}}\mathcal{L}^{(t)}_{\text{global}}.
\end{aligned}
\end{equation}

However, in LMs, $\tfrac{\partial h_t}{\partial y_t}$ does not carry gradient flow, because $y_t$ is obtained as $y_t = W h_t+b$, where $W$ and $b$ denote the output projection matrix and the bias term, respectively, so $h_t$ does not depend on $y_t$. Therefore, we use a heuristic alternative that first differentiates with respect to $h_t$:
\begin{equation}
g_{\text{global}}=\nabla_{h_t}\mathcal{L}^{(t)}_{\text{global}},
\end{equation}
and then project it into the $y_t$ space using the Jacobian
\begin{equation}
\widehat{g}_{\text{hid}}
=J_t\,g_{\text{global}},\qquad
J_t=\tfrac{\partial y_t}{\partial h_t}=W.
\end{equation}
We then use $\widehat{g}_{\text{hid}}$ as a surrogate for $\nabla_{y_t}\mathcal{L}^{(t)}_{\text{global}}$, which empirically works well. Note that in diffusion models, both the latent state $h_t$ and the predicted noise $y_t$ reside in the same dimensional space and are tightly coupled by the scheduler's update rule, so we make a simplifying approximation for diffusion models by treating the required Jacobian as an identity matrix($J_t \approx I$), which is also effective. 
Additionally, \textbf{we greatly reduce} gradient computation by using analytic formulations rather than automatic differentiation(see Appendix \ref{Analytic Gradients for Penalties}).

Finally, to ensure stable magnitude, all penalty gradients $g=\nabla_{y_t}\mathcal{L}^{(t)}_{\,\cdot\,}$ are normalized using
\begin{equation}
\mathrm{Norm}(g)=\frac{g-\mu(g)}{\sqrt{\mathrm{Var}(g)+\varepsilon}},
\end{equation}
where $\mu(g)$ is the mean of $g$ along the last dimension, 
$\mathrm{Var}(g)$ the variance along the same dimension, 
and $\varepsilon$ a small constant for numerical stability.

\begin{figure}[t]
    \centering
    \includegraphics[width=1.0\columnwidth]{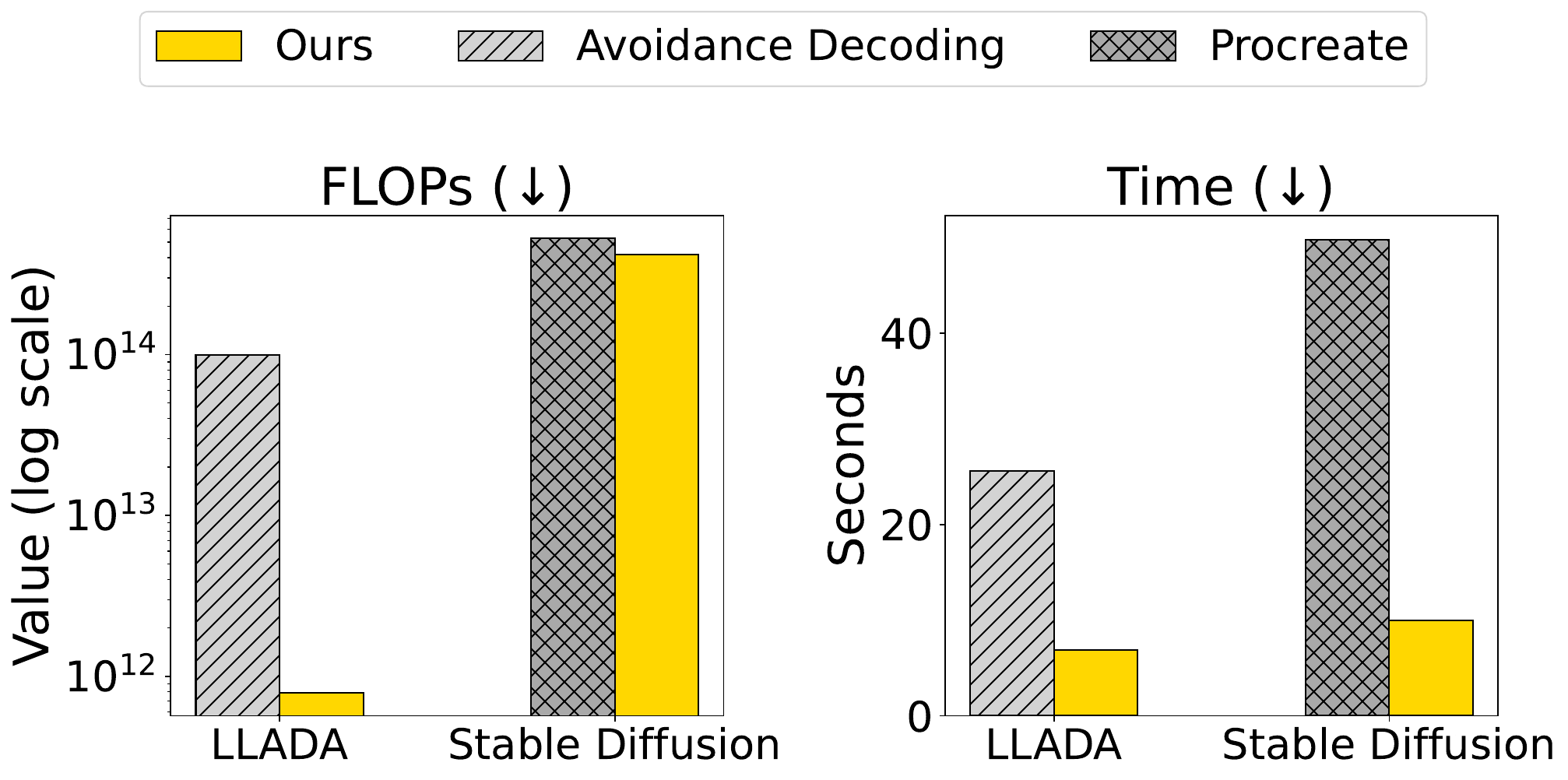}
    \caption{\textbf{FLOPs and Time comparison.}  
    Lower values indicate better performance, and FLOPs are log-scaled due to their large differences.}
    \label{fig:flops_time}
\end{figure}

\begin{table}[t]
\centering
\footnotesize
\setlength{\tabcolsep}{6pt}
\resizebox{\columnwidth}{!}{%
\begin{tabular}{lccc}
\toprule
\header{method} & \header{RougeL($\downarrow$)} & \header{LLM-D($\uparrow$)} & \header{Degen($\downarrow$)} \\
\midrule
\multicolumn{4}{c}{\textbf{Llama-3B}} \\
\midrule
Naive            & 0.257 & 0.293 & \underline{0.005} \\
Top-\emph{k}$^{\dagger}$ & \textit{0.006} & \textit{0.930} & \textit{0.980} \\
Top-\emph{p}     & 0.198 & \underline{0.476} & 0.193 \\
AD     & 0.1265 & \textbf{0.82} & 0.25 \\
Ours$_{global}$  & 0.312 & 0.442 & 0.388 \\
Ours$_{local}$   & \underline{0.212} & 0.391 & \textbf{0.000} \\
\oursrow
Ours             & \textbf{0.093} & \underline{0.680} & 0.565 \\
\midrule
\multicolumn{4}{c}{\textbf{Llada-8B}} \\
\midrule
Naive            & 0.180 & \underline{0.316} & 0.564 \\
Ours$_{global}$  & 0.493 & 0.212 & \underline{0.157} \\
Ours$_{local}$   & \underline{0.082} & 0.278 & \textbf{0.100} \\
\oursrow
Ours             & \textbf{0.067} & \textbf{0.434} & 0.323 \\
\midrule

\header{method} & \header{CLIP($\downarrow$)} & \header{LLM-D($\uparrow$)} & \header{LLM-Q($\uparrow$)} \\
\midrule
\multicolumn{4}{c}{\textbf{Stable Diffusion}} \\
\midrule
Naive           & 0.830 & 0.800 & \textbf{0.720} \\
PC           & \underline{0.218} & 0.76 & \underline{0.684} \\
Ours$_{global}$ & 0.830 & \underline{0.840} & 0.625 \\
Ours$_{local}$  & \textbf{0.143} & \underline{0.840} & 0.355 \\
\oursrow
Ours            & 0.477 & \textbf{0.860} & 0.610 \\
\bottomrule
\end{tabular}%
}
\caption{\textbf{Llama-3B}, \textbf{Llada-8B} for ReedsyPrompts and \textbf{Stable Diffusion} for COCO datasets. AD denotes Avoidance Decoding, and PC denotes Procreate. Best results are in \textbf{bold}, second-best are \underline{underlined}. Top-\emph{k} for Llama-3B is excluded from best/second-best ranking due to extremely high degeneration(0.98). See Appendix \ref{Evaluation results with WritingPrompts} for the full results.}
\label{table_main}
\end{table}

\begin{table}[t]
\centering
\scriptsize
\resizebox{\columnwidth}{!}{%
\begin{tabular}{lcccc}
\toprule
\header{method} & \header{Div($\uparrow$)} & \header{Degen($\uparrow$)} & \header{Crt($\uparrow$)} & \header{Coh($\uparrow$)} \\
\midrule
\multicolumn{5}{c}{\textbf{LlaMa-3B}} \\
\cmidrule(lr){1-5}
Naive         & 1.9 & 1.0 & 1.2 & 1.6 \\
Top-k           & 1.0 & 1.0 & 1.0 & 1.0 \\
Top-p           & 2.1 & 1.0 & 1.3 & 1.7 \\
Ours$_{local}$    & \textbf{4.0} & \textbf{4.1} & \textbf{3.1} & \textbf{3.7} \\
Ours$_{global}$ & 2.8 & 1.6 & 2.6 & \underline{2.8} \\
\oursrow
Ours            & \underline{3.7} & \underline{1.9} & \underline{2.8} & 2.6 \\
\midrule
\multicolumn{5}{c}{\textbf{LlaDa-8B}} \\
\cmidrule(lr){1-5}
Naive         & 2.3 & 1.1 & 1.7 & 1.8 \\
Ours$_{local}$    & \underline{3.1} & \textbf{3.6} & \underline{2.6} & \textbf{3.8} \\
Ours$_{global}$ & 1.1 & \underline{3.2} & 2.0 & 3.3 \\
\oursrow
Ours            & \textbf{3.6} & \underline{3.2} & \textbf{3.3} & \underline{3.5} \\
\midrule
\multicolumn{5}{c}{\textbf{Stable Diffusion}} \\
\cmidrule(lr){1-5}
Naive         & \underline{4.0} & \textbf{3.6} & \textbf{3.8} & \underline{4.0} \\
Ours$_{local}$    & 3.5 & 1.4 & 2.4 & 3.5 \\
Ours$_{global}$ & \textbf{4.25} & \underline{3.0} & \underline{3.4} & \underline{4.0} \\
\oursrow
Ours            & \underline{4.0} & 2.8 & \underline{3.4} & \textbf{4.25} \\
\bottomrule
\end{tabular}%
}
\caption{\textbf{Human evaluation results} across three settings with ReedsyPrompts and COCO. 
Div = Diversity, Degen = Degeneration, Crt = Creativity, Coh = Coherence. Higher is better for all four metrics. Best results are in \textbf{bold}, second-best are \underline{underlined}.}
\label{table4}
\end{table}
\begin{figure}[h] 
    \centering
    \includegraphics[width=0.8\linewidth]{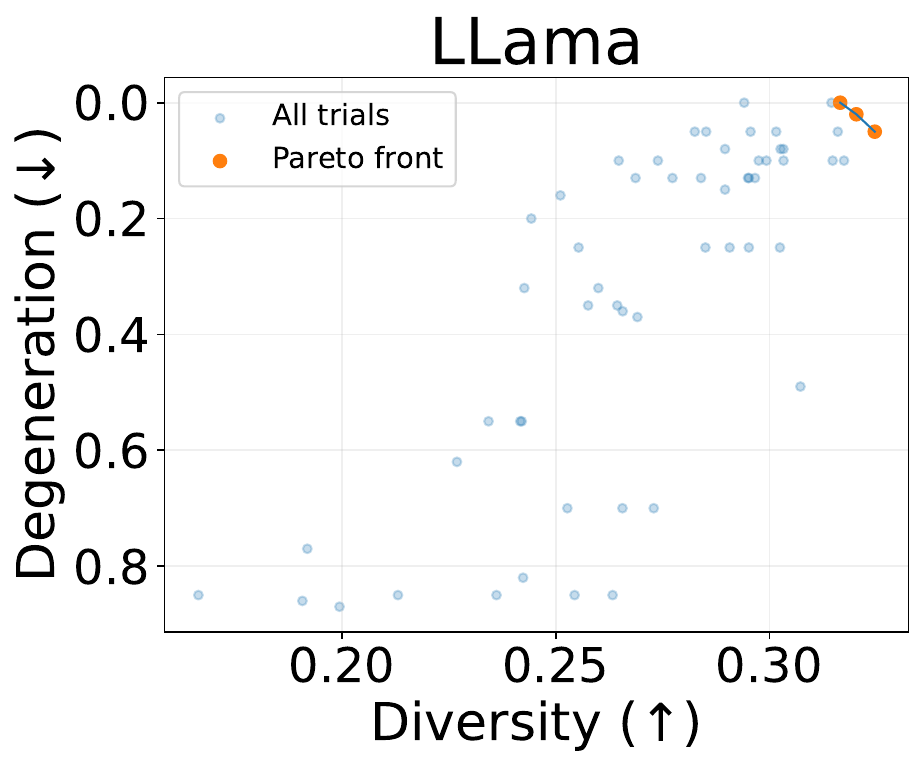}
    \caption{\textbf{Extensive hyperparameter sweeping tests} for LLaMA-3B. We select the best hyperparameter set from the Pareto front points.}
    \label{fig:hyp_6}
\end{figure}

\section{Experiments and Results}

All details on experimental setups such as implementation, datasets, baselines and metrics are in Appendix~\ref{Experimental_Setup}. 

\subsection{Flops and Time Comparison}
\label{flops}
As shown in Figure \ref{fig:flops_time}, the previous state-of-the-art textual generation method, \emph{Avoidance Decoding}, incurs approximately 126 times more FLOPs and requires 3.73 times longer runtime than our method. In image generation, the prior state-of-the-art method, \emph{Procreate}, requires about 1.25 times more FLOPs and 5.01 times longer runtime.Therefore, we conclude that our method achieves substantially better computational efficiency by relying on simple gradient computation rather than token-wise penalty operations or external model calls.

\subsection{Automatic Evaluation}
As shown in Table \ref{table_main}, in the LLaMA-3B test, it achieves strong diversity even when compared with the strong baseline, \emph{Avoidance Decoding}, but exhibits relatively higher degeneration due to the inherent trade-off; nevertheless, under our strict evaluation criteria, the outputs remain acceptably readable (see Appendix \ref{Qualitative Analysis of Degeneration}). For the LlaDA-8B diffusion language model, it maintains balanced performance with superior diversity compared to high-temperature baselines. In diffusion-based image generation, our method achieves the highest LLM-Diversity and competitive CLIP scores, with an acceptable trade-off between diversity and quality, even when compared with the strong baseline, \emph{Procreate}. Furthermore, we conduct large-scale model experiments which shows enhanced diversity and degeneration scores (see Appendix \ref{large_scale_model_adaptation}).  We therefore conclude that our method enhances multi-branch diversity successfully than either ablated versions or other state-of-the-art methods. 

\subsection{Human Evaluation}
\label{Human Evaluation}

See Appendix \ref{Human Evaluation Details} for details including annotator information, rubrics and agreements. 
As shown in Table \ref{table4}, our method consistently ranked among the top across all models: second-best overall for LLaMA, best in diversity and creativity for LLada, and best in coherence with strong diversity and creativity for Stable Diffusion. These results demonstrate that the combined penalty in UAG effectively enhances creativity, diversity, and coherence across different models, while maintaining acceptable degeneration compared to naive or ablated versions.

\subsection{Searching for the Best Hyperparameters}
To identify the best trade-off between diversity and degeneration, we perform extensive hyperparameter sweeping tests \textbf{for all baselines}. As a result, we find the optimal hyperparameters that achieve the lowest degeneration while yielding the highest diversity, as shown in Figure \ref{fig:hyp_6}. Since diversity and degeneration are positively correlated during sweeping, this confirms that our hyperparameter search is a right approach to determine the optimal balance. See Appendix \ref{Extensive Experiments for Best Hyper-parameters} for additional details and other sweeping experiment results.

\section{Conclusion}
We introduce UAG, an efficient framework that enhances multi-branch diversity across autoregressive and diffusion-based generative models using gradient-based  penalties. UAG achieves top-ranked strong diversity with acceptable degeneration, while significantly reducing FLOPs and runtime compared to prior state-of-the-art methods.
\section*{Limitations}

Our framework is not applicable to generative models which lie outside the scope of the Universal Generation Process, such as GANs. Developing methods that can be universally applied across all classes of generative models still remains as a challenge for future research.

\section*{Ethical Considerations}
This work primarily focuses on improving the diversity of generative models and therefore does not directly raise ethical concerns. However, we acknowledge that excessive pursuit of diversity could potentially violate ethical boundaries. Future research should consider safety and ethical safeguards to ensure that enhanced diversity does not compromise responsible use.

\section*{Acknowledgement}
We thank anonymous reviewers for their constructive and insightful comments. K. Jung is with ASRI, Seoul National University, Korea. This work was partly supported by Institute of Information \& communications Technology Planning \& Evaluation (IITP) grant funded by the Korea government(MSIT) (No.RS-2022-II220184, Development and Study of AI Technologies to Inexpensively Conform to Evolving Policy on Ethics), Institute of Information \& Communications Technology Planning \& Evaluation(IITP)-ITRC(Information Technology Research Center) grant funded by the Korea government(MSIT)(IITP-2025-RS-2024- 00437633, 30\%)], and the National Research Foundation of Korea(NRF) grant funded by the Korea goverment(MSIT) (RS-2025-02263628).

\bibliography{acl_latex}

\appendix

\section{Human Evaluation Details}
\label{Human Evaluation Details}
\subsection{Annotators Details}
We recruited five graduate and undergraduate students fluent in English. The recruited annotators were provided with a detailed description of task definitions, instructions, and samples of each model. Also, all applicants were informed that their annotations would be used for academic purposes and would be published in paper material through the recruitment announcement and instructions. \\ Each of the five annotators was given six samples—each consisting of five outputs from all baselines—and answered four questions for each sample.
 For the payment of the annotators, the co-authors conducted annotations for 6 hours first to estimate the average number of annotations that could be completed in the same time. Based on this estimation, a rate of 0.5 dollars per example was established to ensure that the annotators would be paid at least the minimum wage.

\subsection{Rubrics}
The detailed rubrics for human evaluation are in Table \ref{fig:human_rubric},\ref{fig:human_rubric_image}.

\begin{figure*}[t]
\begin{tcolorbox}[
    width=\textwidth,       
    sharp corners,          
    colback=white,          
    colframe=black,         
    boxrule=0.5pt,          
    left=10pt, right=10pt,  
    top=8pt, bottom=8pt     
]
\textbf{Question 1. Diversity} \\
To what extent are the generated stories fundamentally different from each other, beyond merely changing character names or surface-level details? (1–5)

\vspace{0.5em}
\textbf{Question 2. Degeneration} \\
To what extent are the sentences fluent, grammatical, and lexically natural, without exhibiting broken syntax or incoherent word usage? (1 = very poor, 5 = excellent)

\vspace{0.5em}
\textbf{Question 3. Creativity} \\
To what extent are the stories novel, engaging, and imaginative, rather than conventional or obvious? (1 = very poor, 5 = excellent)

\vspace{0.5em}
\textbf{Question 4. Coherence} \\
To what extent does each story develop in a consistent and logically connected manner, maintaining narrative coherence throughout? (1 = very poor, 5 = excellent)
\end{tcolorbox}
\caption{Human Evaluation Rubric for Measuring Textual Diversity, Degeneration, Creativity, and Coherence.}
\label{fig:human_rubric}
\end{figure*}

\begin{figure*}[t]
\begin{tcolorbox}[
    width=\textwidth,       
    sharp corners,          
    colback=white,          
    colframe=black,         
    boxrule=0.5pt,          
    left=10pt, right=10pt,  
    top=8pt, bottom=8pt     
]
Diversity: Are the stories fundamentally different from each other?(1 = very poor, 5 = excellent)

Degeneration: Are the sentences natural and fluent, without being broken or degraded?(1 = very poor, 5 = excellent)

Creativity: Are the stories non-trivial, engaging, and imaginative rather than obvious?(1 = very poor, 5 = excellent)

Coherence: Are the stories consistent with the given prompt?(1 = very poor, 5 = excellent)
\end{tcolorbox}
\caption{Human Evaluation Rubric for Measuring Image Diversity, Degeneration, Creativity, and Coherence.}
\label{fig:human_rubric_image}
\end{figure*}

\subsection{Agreements}
To assess annotator agreement, we computed both Kendall’s coefficient of concordance(W) and intraclass correlation coefficients(ICC) across five raters. Kendall’s W reached 0.61, suggesting moderate to substantial agreement among annotators. The single‐measure ICC(1,1) was 0.49, indicating moderate reliability. Taken together, these results show that annotator ratings exhibit a reasonable level of consistency, though not uniformly high across all items.

\section{Mathematical Motivation of UAG}
\label{Mathematical Intuition of UAG}

We recall the standard smoothness assumption: a differentiable function 
$\mathcal{L}:\mathbb{R}^d \to \mathbb{R}$ is said to have an $L$-Lipschitz 
continuous gradient if for all $x,y \in \mathbb{R}^d$,
\[
\|\nabla \mathcal{L}(x)-\nabla \mathcal{L}(y)\|\;\le\;L\|x-y\|.
\]
This condition is satisfied by all similarity functions we employ in 
$\mathcal{L}^{(t)}_{\text{UAG}}$: dot-product similarity(used in 
autoregressive and diffusion language models), cosine similarity 
(used in diffusion latent penalties), and CLIP-based similarity 
(used in diffusion image penalties, where CLIP encoders are smooth neural 
networks). Hence $\nabla \mathcal{L}^{(t)}_{\text{UAG}}$ is 
$L$-Lipschitz for some constant $L>0$.

Under this assumption, a first-order Taylor expansion yields
\begin{align*}
\mathcal{L}^{(t)}_{\text{UAG}}(\hat y_t)
&\le \mathcal{L}^{(t)}_{\text{UAG}}(y_t) \notag \\
&\quad - \eta \,\bigl\|\nabla_{y_t}\mathcal{L}^{(t)}_{\text{UAG}}\bigr\|^2
+ O(\eta^2).
\end{align*}
so for sufficiently small step size $\eta$ the penalty strictly decreases, 
thereby reducing similarity to reference bank items. More generally, the 
smoothness inequality implies that for any update 
$y_t^{+}=y_t-\eta g$ with $g=\nabla_{y_t}\mathcal{L}^{(t)}_{\text{UAG}}$, 
\[
\mathcal{L}^{(t)}_{\text{UAG}}(y_t^{+}) 
\le \mathcal{L}^{(t)}_{\text{UAG}}(y_t)
-\eta\Bigl(1-\tfrac{L\eta}{2}\Bigr)\|g\|^2,
\]
\[
\qquad\text{for }\;\eta<2/L.
\]
Thus, the update guarantees a monotonic decrease in the UAG penalty, 
ensuring that each step progressively increases dissimilarity from past 
generations and thereby promotes diversity across branches.

\section{Analytic Gradients for Penalties}
\label{Analytic Gradients for Penalties}
We avoid \texttt{torch.autograd.grad} over the model graph as much as possible. Only CLIP loss traverses the external differentiable path(VAE$\to$CLIP), thus uses \texttt{torch.autograd}.

\paragraph{Repulsion loss(logit-based).}
Let $\mathcal{B}_t^{\text{out}}=\{q_1,\dots,q_N\}$ with each $q_r$ a reference distribution.
Then as $\phi(y_t)=p_t=\mathrm{softmax}(y_t)$, 
\[
\nabla_{y_t}\mathcal{L}^{(t)}_{\text{rep}}
=\frac{1}{N}\sum_{r=1}^N \bigl(p_t \odot q_r -(p_t^\top q_r)\,p_t\bigr).
\]

\paragraph{Hidden-state loss.}

If $b^*=\arg\max_{b\in\mathcal{B}^{\text{hid}}_t}\langle h_t,b\rangle$, then
\[
\nabla_{y_t}\mathcal{L}^{(t)}_{\text{hid}} = W b^*.
\]

\paragraph{Latent cosine loss(Diffusion).}
Flatten the latent $z_t \in \mathbb{R}^M$, and let $\{y_j\}$ be past latents.  
Define $\cos(z_t,y) = \tfrac{\langle z_t,y\rangle}{\|z_t\|\|y\|}$ and $y^*=\arg\max_j \cos(z_t,y_j)$. Then
\[
\nabla_{z_t}\cos(z_t,y^*)
= \frac{y^*}{\|z_t\|\|y^*\|} - \frac{\cos(z_t,y^*)}{\|z_t\|^2}\,z_t.
\]

\paragraph{Noise(CLIP) loss.}
For the CLIP-guided objective that depends on VAE decoding and CLIP embeddings, 
we rely on automatic differentiation \texttt{torch.autograd} to obtain $\nabla_{z_t}\mathcal{L}^{(t)}_{\text{noise}}$.

\section{Experimental Setup}
\label{Experimental_Setup}
\subsection{Datasets}
We use the {ReedsyPrompts} \cite{park2024longstory} and {WritingPrompts} \cite{fan2018hierarchical} datasets for story generation, and {COCO} \cite{lin2014microsoft} for image generation. For each dataset, we take the first 20 prompts or annotations and generate 15 multi-branch stories or images from the same prompt or annotation, while  avoiding previously generated outputs by our method.

\subsection{Implementation Details}

All training and evaluation are performed on a single NVIDIA A100 GPU(40~GB memory). For all LLM-based qualitative evaluation of \emph{diversity} and \emph{quality}, we employ OpenAI's \texttt{GPT-4.1-2025-04-14} \cite{openai2025gpt4_1}, following the rubrics in Appendix \ref{LLM Evaluation Details}. To identify optimal hyperparameters, we conduct $\sim$300 runs with a dense sweep over the parameter space, analyzing both diversity and degeneration metrics(see Appendix \ref{Extensive Experiments for Best Hyper-parameters} for details). The resulting best configurations are:
\begin{itemize}
  \item \textbf{LLada-8B}: $\alpha=1.766$, $\beta=1.077$, $L_{0}=38$, $\delta=0.8024$
  \item \textbf{LLaMA-3B}: $\alpha=0.3395$, $\beta=1.3339$, $L_{0}=5$, $\delta=0.5479$
  \item \textbf{Stable Diffusion v1.5}: $\alpha=0.0579$, $\beta=0.0208$, $L_{0}=51$, $\delta=1.8268$
\end{itemize}
Additional details on the settings of the ablation versions are provided in Appendix \ref{Extensive Experiments for Best Hyper-parameters}. Unless otherwise noted, text output length is fixed to 200 tokens. For Stable Diffusion v1.5 we use 50 diffusion steps, and for LLada-8B we use 200 steps. For $\mathrm{sim}(\cdot,\cdot)$, both autoregressive and diffusion LMs use dot-product for $L_{local}$ and $L_{global}$, whereas diffusion image models use cosine similarity for $L_{local}$ and CLIP-based similarity on VAE-decoded images for $L_{global}$. Additionally, all Language models project hidden states gradient to logit space via $W$.

\subsection{Baselines}
Our baselines include ablated versions of our method, Ours$_{\text{local}}$ and Ours$_{\text{global}}$, as well as standard sampling methods such as {top-}$k$, {top-}$p$, and {naive} sampling with high temperature. Ours$_{\text{local}}$ applies only the model local similarity penalty, while Ours$_{\text{global}}$ applies only the global similarity penalty. We also compare the runtime and FLOPs of our method with \emph{Avoidance Decoding}, the current state-of-the-art approach for diverse multi-branch story generation.

\subsection{Metrics across Models}
We conduct experiments with our method on three types of generative models.

First, we evaluate our method on an autoregressive language model, the {LLaMA-3B}.
Second, we evaluate our method on a diffusion-based language model, the {LLada-3B}. For both models, to measure the diversity, we use $n$-gram repetition-based metrics such as BLEU, ROUGE-L, and METEOR, as well as cosine similarity between sentence embeddings obtained from Sentence-BERT. In addition, we employ OpenAI’s GPT-4.1 to assess the overall diversity(LLM-D) and the degeneration(Degen) of the samples.
Third, we evaluate a diffusion-based image model by applying UAG to {Stable Diffusion v1.5}. For diversity evaluation, we calculate the CLIP-Score between generated images to assess pairwise similarity, and further employ OpenAI’s GPT-4.1 to evaluate overall diversity(LLM-D) and quality(LLM-Q). Finally, we conduct a human evaluation to assess creativity, diversity, degeneration, and coherence across all types of generation models. Additionally,  section \ref{flops} shows significantly lower \emph{runtime} and \emph{computational cost} of our method than previous state-of-the-art methods, such as Avoidance Decoding \cite{park2025avoidance} and Procreate \cite{lu2024procreate}. Other additional experimental results  are provided in Appendix \ref{Evaluation results with WritingPrompts}, which consistently align with the results here.

\begin{table}[h] 
\centering
\begin{tcolorbox}[
    colback=gray!5!white, 
    colframe=black!75!white, 
    sharp corners, 
    fontupper=\itshape, 
    title=The Most Degenerated Story 
]
"It started as an accident.  She woke up one morning and found her car parked in front of her house, 100 yards from where it was supposedly parked last night!  She brushed it off as a fluke, but the next morning, her car was still parked 100 yards away from where it was supposed to be parked!  She freaked out!  She swore she swore she swore she hadn't moved the parking spot!  The next morning! Her car was still parked 100 yards away from her parking spot!  She freaked out!  She swore she swore she swore she hadn't moved the parking spot! The next morning! Her car was still parked 100 yards away from her parking spot.  She freaked out!  She swore she swore she swore she hadn't moved the parking spot! And there she goes! A character who keeps ending up in the same place!"

\end{tcolorbox}
\caption{The most degenerated sample in the LLaMA-3B experiments.} 
\label{sample}
\end{table}

\section{Scheduling Methods Ablation Studies}
\label{Scheduling Methods Ablation Studies}

We additionally experimented with two ablation version of scheduling method, the constant and linear scheduling methods. The constant scheduling method keeps the weights of the two penalties fixed throughout generation, whereas the linear scheduling method adjusts the weights linearly while preserving a constant sum between them. As table \ref{tab:llm_results}, \ref{tab:diffusion_lm_results}, \ref{tab:diffusion_image_results}, logistic scheduling achieves the highest diversity. 

\section{Qualitative Analysis of Degeneration}
\label{Qualitative Analysis of Degeneration}
The relatively high degeneration score of our method is expected because of our highly strict degeneration evaluation rubric (in both automatic and human evaluation). For reference, table \ref{sample} is the most severely degenerated output at LLaMA-3B’s score of 0.565, which still exhibit storytelling ability.

\begin{table*}[t]
\centering
\footnotesize

\begin{tabular}{lcccccc}
\toprule
Method & BLEU $\downarrow$ & Rouge-L $\downarrow$ & METEOR $\downarrow$ & SBERT $\downarrow$ & LLM-D $\uparrow$ & Degen $\downarrow$ \\ \midrule
logistic (ours) & \textbf{0.016} & \textbf{0.093} & \textbf{0.1364} & \textbf{0.430} & \textbf{0.6794} & 0.5645 \\
const           & 0.0776          & 0.2018          & 0.2844          & 0.6572         & 0.44           & \textbf{0.0}   \\
linear          & 0.1059          & 0.2246          & 0.2973          & 0.6644         & 0.32           & \textbf{0.0}   \\ \bottomrule
\end{tabular}
\caption{Llama-3B Logitstic Scheduling Ablation Study.}
\label{tab:llm_results}
\end{table*}


\begin{table*}[h]
\centering
\footnotesize
\begin{tabular}{lcccccc}
\toprule
Method & BLEU $\downarrow$ & Rouge-L $\downarrow$ & METEOR $\downarrow$ & SBERT $\downarrow$ & LLM-D $\uparrow$ & Degen $\downarrow$ \\ \midrule
logistic (ours) & 0.0679          & \textbf{0.1811} & \textbf{0.2270} & \textbf{0.5242} & \textbf{0.434} & 0.3235 \\
const           & \textbf{0.0533} & 0.1869          & 0.2512          & 0.5839          & 0.29           & 0.035  \\
linear          & 0.0927          & 0.2277          & 0.2704          & 0.6445          & 0.33           & \textbf{0.0}   \\ \bottomrule
\end{tabular}

\caption{Llada-8B Logitstic Scheduling Ablation Study.}
\label{tab:diffusion_lm_results}
\end{table*}

\begin{table}[h]
\centering
\resizebox{\columnwidth}{!}{%
\begin{tabular}{lccc}
\toprule
Method & CLIP $\downarrow$ & LLM-D $\uparrow$ & LLM-Q $\uparrow$ \\ \midrule
logistic (ours) & \textbf{0.477} & \textbf{0.860} & 0.610          \\
const           & 0.8759         & 0.750          & 0.6625         \\
linear          & 0.9089         & 0.750          & \textbf{0.7375} \\ \bottomrule
\end{tabular}
}
\caption{Stable Diffusion Logitstic Scheduling Ablation Study.}
\label{tab:diffusion_image_results}
\end{table}

\section{Large Scale Model Adaptation}
\label{large_scale_model_adaptation}
We additionally performed large-scale experiments on LLaMA-70B \cite{dubey2024llama}, stable-diffusion-3.5-large (8B) \cite{stable_diffusion_3_5}, under the same settings as in the main experiments of the paper. These include full hyperparameter sweeps; the best results are shown in table \ref{tab:llama_70b}, \ref{tab:sd_large}. The results show that as model size increases, degeneration is better suppressed, proving wide applicability.

\begin{table*}[t]
\centering
\footnotesize
\begin{tabular}{lcccccc}
\toprule
Method & BLEU $\downarrow$ & ROUGE-L $\downarrow$ & METEOR $\downarrow$ & SBERT $\downarrow$ & Degen $\downarrow$ & LLM-D $\uparrow$ \\ \midrule
ours              & \textbf{0.0788} & 0.2010          & 0.2801          & 0.6267          & 0.0             & 0.52 \\
Naive (high temp)  & 0.1069          & 0.2281          & 0.3005          & 0.6768          & \textbf{0.0}    & 0.34 \\ \bottomrule
\end{tabular}
\caption{LLaMA-70B experiment results. }
\label{tab:llama_70b}
\end{table*}

\begin{table}[h]
\centering

\resizebox{\columnwidth}{!}{%
\begin{tabular}{lccc}
\toprule
Method & CLIP $\downarrow$ & LLM-D $\uparrow$ & LLM-Q $\uparrow$ \\ \midrule
ours                & \textbf{0.9023} & \textbf{0.700} & 0.7313          \\
Naive (random seed) & 0.9090          & 0.450          & \textbf{0.7438} \\ \bottomrule
\end{tabular}
}
\caption{Stable Diffusion Large (8B) Results.}
\label{tab:sd_large}
\end{table}

\section{Full Evaluation results}
\label{Evaluation results with WritingPrompts}
The full evaluation results using the same metrics on different datasets(ReedsyPrompts and WritingPrompts) are reported in Tables \ref{table1}, \ref{table2}, \ref{table1_writing}, and \ref{table2_writing}. Additionally, we conduct the same human evaluation of the section \ref{Human Evaluation} on WritingPrompts, as reported in Table \ref{table3_writing}.

\begin{table*}[t] \centering \setlength{\tabcolsep}{7pt} \resizebox{\textwidth}{!}{%
\begin{tabular}{lcccccc} \toprule \header{method} & \header{BLEU($\downarrow$)} & \header{RougeL($\downarrow$)} & \header{METEOR($\downarrow$)} & \header{Sent-Sim($\downarrow$)} & \header{LLM-D($\uparrow$)} & \header{Degen($\downarrow$)} \\ 

\midrule Naive & 0.0840 & 0.2410 & 0.2570 & 0.4700 & 0.2930 & \underline{0.0050} \\
Top-\emph{k}$^{\dagger}$ & \textit{0.0036} & \textit{0.0059} & \textit{0.0180} & \textit{0.2457} & \textit{0.9300} & \textit{0.9800} \\
Top-\emph{p} & 0.0573 & 0.1978 & 0.2026 & \textbf{0.4041} & 0.4755 & 0.1930 \\
Avoidance Decoding & \textbf{0.0112} & \underline{0.1265} & \underline{0.1483} & \underline{0.4136} & \textbf{0.82} & 0.25 \\
\midrule 
Ours$_{global}$ & 0.2540 & 0.3119 & 0.3190 & 0.6570 & 0.4415 & 0.3875 \\
Ours$_{local}$ & 0.1014 & 0.2120 & 0.2943 & 0.6100 & 0.3905 & \textbf{0.0000} \\ 
\oursrow 
Ours &  \underline{0.0165} & \textbf{0.0930} & \textbf{0.1360} & 0.4300 & \underline{0.6795} & 0.5645 \\ 

\bottomrule \end{tabular}} \caption{\textbf{Llama-3B} with ReedsyPrompts multi-branch story generation results. Lower is better for BLEU, RougeL, METEOR, Sent-Sim, Time, and Degen; higher is better for LLM-D. Best results are highlighted in \textbf{bold} and second-best results are \underline{underlined}. Top-\emph{k} achieved the best scores on several metrics but is excluded from best/second-best ranking due to extremely high degeneration(0.98).} \label{table1} \end{table*} 

\begin{table*}[t] \centering \setlength{\tabcolsep}{7pt} \resizebox{\textwidth}{!}{%
\begin{tabular}{lcccccc} \toprule \header{method} & \header{BLEU($\downarrow$)} & \header{RougeL($\downarrow$)} & \header{METEOR($\downarrow$)} & \header{Sent-Sim($\downarrow$)} & \header{LLM-D($\uparrow$)} & \header{Degen($\downarrow$)} \\ \midrule Naive & 0.180 & 0.344 & 0.356 & \textbf{0.517} & \underline{0.316} & 0.564 \\ Ours$_{global}$ & 0.493 & 0.552 & 0.565 & 0.778 & 0.212 & \underline{0.157} \\ Ours$_{local}$ & \underline{0.082} & \underline{0.213} & \underline{0.269} & 0.625 & 0.278 & \textbf{0.100} \\ \oursrow Ours & \textbf{0.067} & \textbf{0.181} & \textbf{0.227} & \underline{0.524} & \textbf{0.434} & 0.323 \\ \bottomrule \end{tabular}}
\caption{\textbf{Llada-8B} with ReedsyPrompts multi-branch story generation results. Lower is better for BLEU, RougeL, METEOR, Cos, and Degen; higher is better for LLM-D. Best results are highlighted in \textbf{bold} and second-best results are \underline{underlined}.} \label{table2} \end{table*}
\begin{table*}[t]
\centering
\setlength{\tabcolsep}{7pt}
\resizebox{\textwidth}{!}{%
\begin{tabular}{lcccccc}
\toprule
\header{method} & \header{BLEU($\downarrow$)} & \header{RougeL($\downarrow$)} & \header{METEOR($\downarrow$)} & \header{Sent-Sim($\downarrow$)} & \header{LLM-D($\uparrow$)} & \header{Degen($\downarrow$)} \\
\midrule
Naive          & 0.0646 & 0.2402 & 0.2410 & 0.4705 & 0.3790 & \underline{0.0175} \\
Top-\emph{k}$^{\dagger}$ & \textit{0.0036} & \textit{0.0095} & \textit{0.0173} & \textit{0.2781} & \textit{0.9300} & \textit{0.9800} \\
Top-\emph{p}   &0.0388 & 0.1878 & 0.1850 & \underline{0.3851} & 0.5190 & 0.2395 \\
Avoidance Decoding & \textbf{0.0130} & \underline{0.1262} & \underline{0.1632} & \textbf{0.3708} & \textbf{0.7965} & 0.27 \\
\midrule
Ours$_{global}$   & 0.2192 & 0.2898 & 0.2957 & 0.6118 & 0.4510 & 0.4690 \\
Ours$_{local}$      & 0.1109 & 0.2244 & 0.3065 & 0.6575 & 0.3770 & \textbf{0.0075} \\
\oursrow
Ours              & \underline{0.0170} & \textbf{0.1033} & \textbf{0.1488} & 0.4699 & \underline{0.6805} & 0.4875 \\
\bottomrule
\end{tabular}
}
\caption{\textbf{Llama-3B} with WritingPrompts multi-branch story generation results. Lower is better for BLEU, RougeL, METEOR, Sent-Sim, Time, and Degen; higher is better for LLM-D. Best results are highlighted in \textbf{bold} and second-best results are \underline{underlined}. Top-\emph{k} achieved the best scores on several metrics but is excluded from best/second-best ranking due to extremely high degeneration(0.98).}
\label{table1_writing}
\end{table*}

\begin{table*}[t]
\centering
\setlength{\tabcolsep}{7pt}
\resizebox{\textwidth}{!}{%
\begin{tabular}{lcccccc}
\toprule
\header{method} & \header{BLEU($\downarrow$)} & \header{RougeL($\downarrow$)} & \header{METEOR($\downarrow$)} & \header{Cos($\downarrow$)} & \header{LLM-D($\uparrow$)} & \header{Degen($\downarrow$)} \\
\midrule
High temperature & 0.281 & 0.479 & 0.444 & \textbf{0.453} & 0.178 & 0.398 \\
Ours$_{global}$  & 0.586 & 0.638 & 0.631 & 0.829 & 0.193 & \underline{0.258} \\
Ours$_{local}$     & \underline{0.072} & \underline{0.195} & \underline{0.233} & 0.593 & \underline{0.353} & 0.320 \\
\oursrow
Ours             & \textbf{0.069} & \textbf{0.186} & \textbf{0.205} & \underline{0.530} & \textbf{0.435} & 0.506 \\
\bottomrule
\end{tabular}
}
\caption{\textbf{LLADA-8B} with WritingPrompts multi-branch story generation results. Lower is better for BLEU, RougeL, METEOR, Cos, and Degen; higher is better for LLM-D. Best results are highlighted in \textbf{bold} and second-best results are \underline{underlined}.}
\label{table2_writing}
\end{table*}

\begin{table}[t]
\centering
\scriptsize
\resizebox{\columnwidth}{!}{%
\begin{tabular}{lcccc}
\toprule
\header{method} & \header{Div($\uparrow$)} & \header{Degen($\uparrow$)} & \header{Crt($\uparrow$)} & \header{Coh($\uparrow$)} \\
\midrule
\multicolumn{5}{c}{\textbf{LLaMA}} \\
\cmidrule(lr){1-5}
Ours$_{global}$ & 3.0 & 2.0 & 3.0 & 2.8 \\
Ours$_{local}$    & \textbf{3.7} & \textbf{3.7} & \textbf{3.6} & \textbf{4.0} \\
Naive           & 2.3 & 1.4 & 1.4 & 1.7 \\
Top-k           & 1.0 & 1.0 & 1.0 & 1.0 \\
Top-p           & 2.1 & 1.5 & 1.6 & 1.8 \\
\oursrow
Ours            & \textbf{3.7} & \underline{2.1} & \underline{3.3} & \underline{3.3} \\
\midrule
\multicolumn{5}{c}{\textbf{LLaDA}} \\
\cmidrule(lr){1-5}
Ours$_{global}$ & 1.0 & \underline{2.3} & 1.7 & 2.6 \\
Ours$_{local}$    & \textbf{3.4} & \textbf{3.2} & \textbf{3.3} & \underline{3.3} \\
Naive           & 1.6 & 1.0 & 1.2 & 1.0 \\
\oursrow
Ours            & \textbf{3.4} & 2.2 & \textbf{3.3} & \textbf{3.5} \\
\bottomrule
\end{tabular}%
}
\caption{\textbf{Human evaluation results with WritingPrompts.} 
Div = Diversity, Degen = Degeneration, Crt = Creativity, Coh = Coherence. 
Higher is better for all metrics. Best results are highlighted in \textbf{bold}, and second-best in \underline{underlined}.}
\label{table3_writing}
\end{table}

\section{Extensive Experiments for Best Hyper-parameters}
\label{Extensive Experiments for Best Hyper-parameters}
\subsection{Autoregressive Language Model Test}
We conduct a total of 33, 33, 33, and 200 experiments to determine the best hyperparameter settings for Ours$_{local}$, Ours$_{global}$, Naive, and Ours, respectively. The ablation and Naive versions involve only a single variable—$\alpha$ for Ours$_{local}$, $\beta$ for Ours$_{global}$, and temperature for Naive—so we run far fewer experiments for them. For top-$k$ and top-$p$ tests, we set $k=20$ and $p=0.9$. In the naive image generation tests, we use a different random seed for each sample. For all textual experiments, we exclusively use the ReedsyPrompts dataset. Results with severe degeneration(Degen $>$ 0.9) are omitted from the figures.

\begin{figure}[h] 
    \centering
    \includegraphics[width=0.8\linewidth]{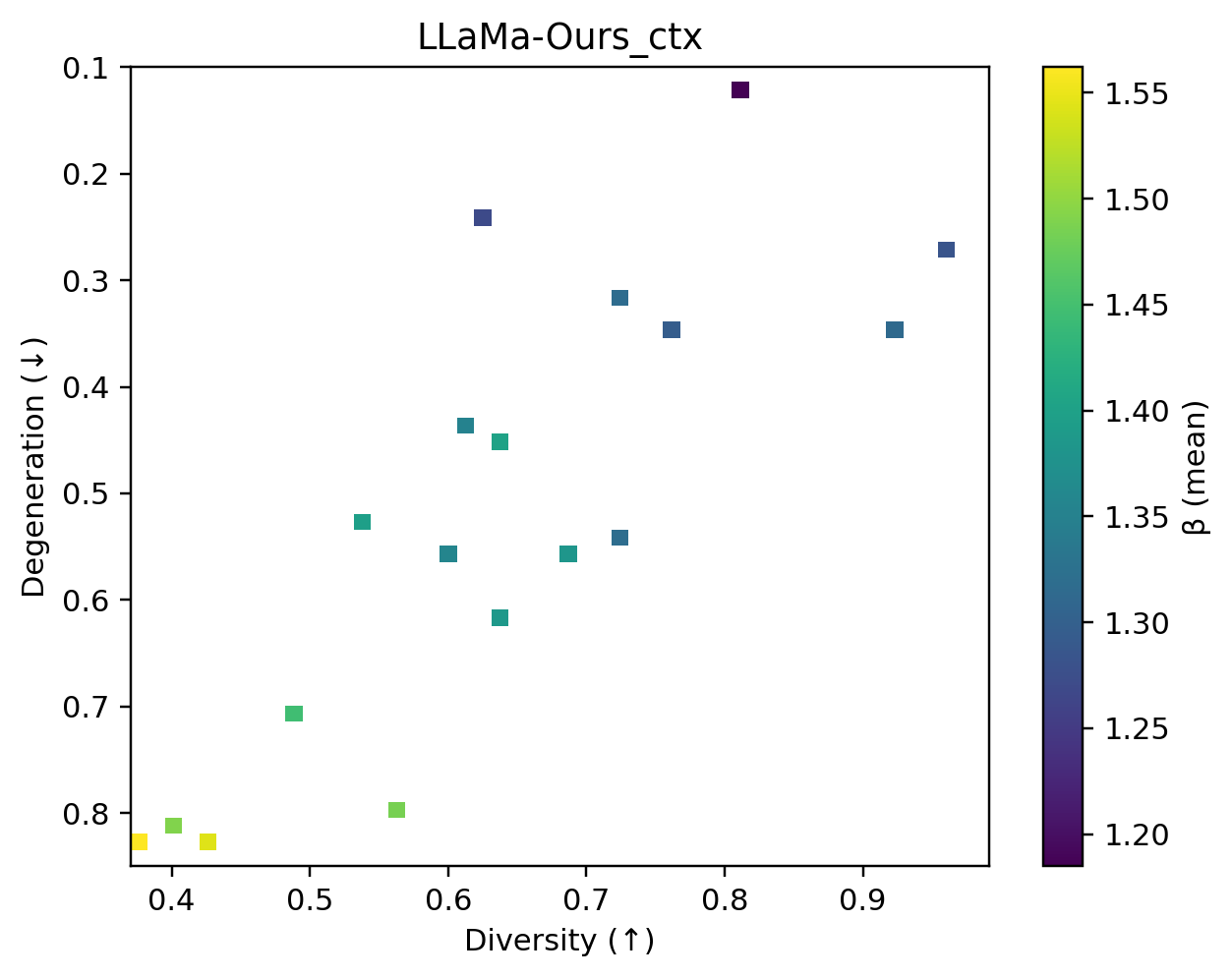}
    \caption{Hyperparameter sweep results for Ours$_{global}$ on the LlaMA-3B.}
    \label{fig:hyp_1}
\end{figure}

\begin{figure}[h] 
    \centering
    \includegraphics[width=0.8\linewidth]{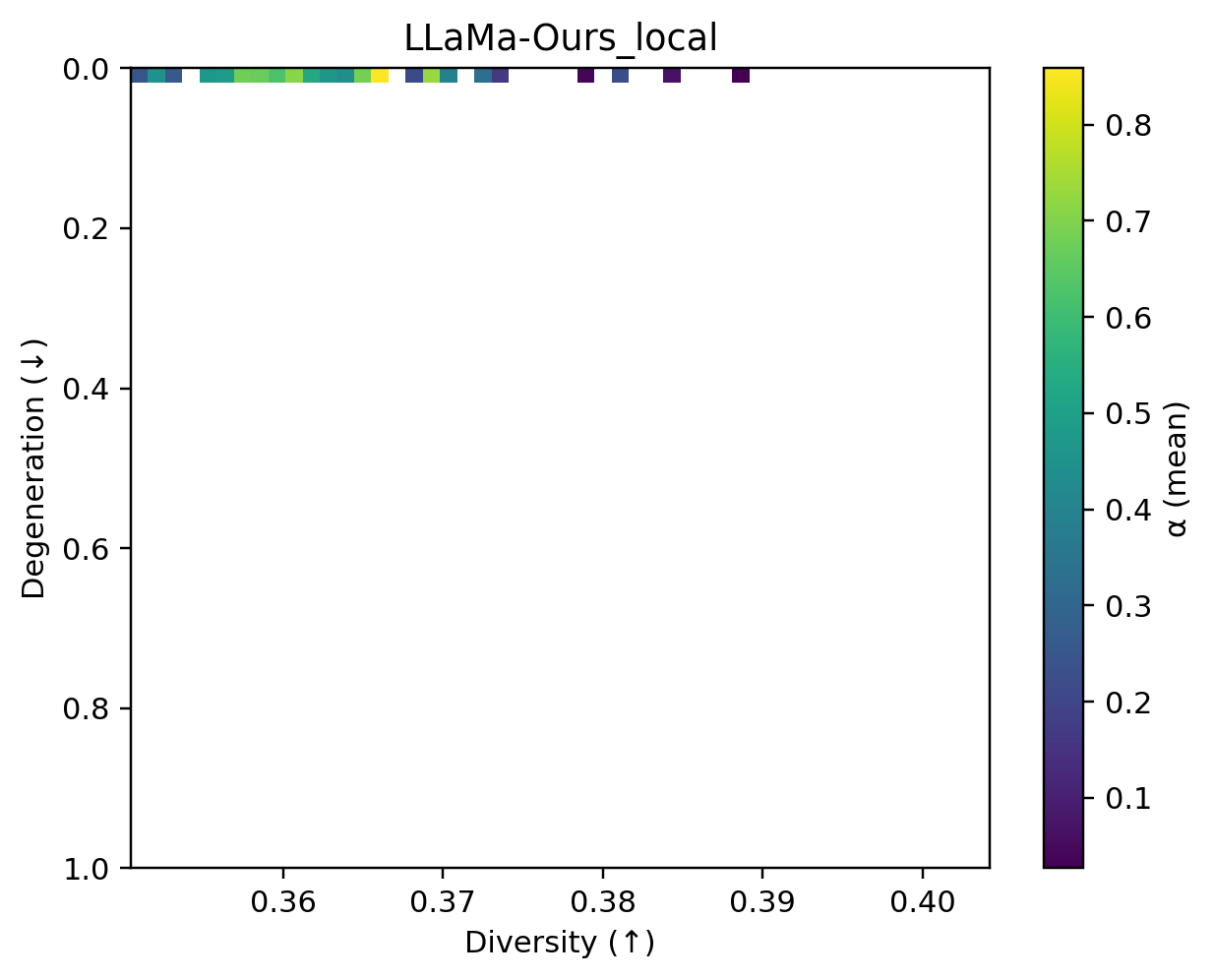}
    \caption{Hyperparameter sweep results for Ours$_{local}$ on the LlaMA-3B.}
    \label{fig:hyp_2}
\end{figure}

\begin{figure}[h] 
    \centering
    \includegraphics[width=0.8\linewidth]{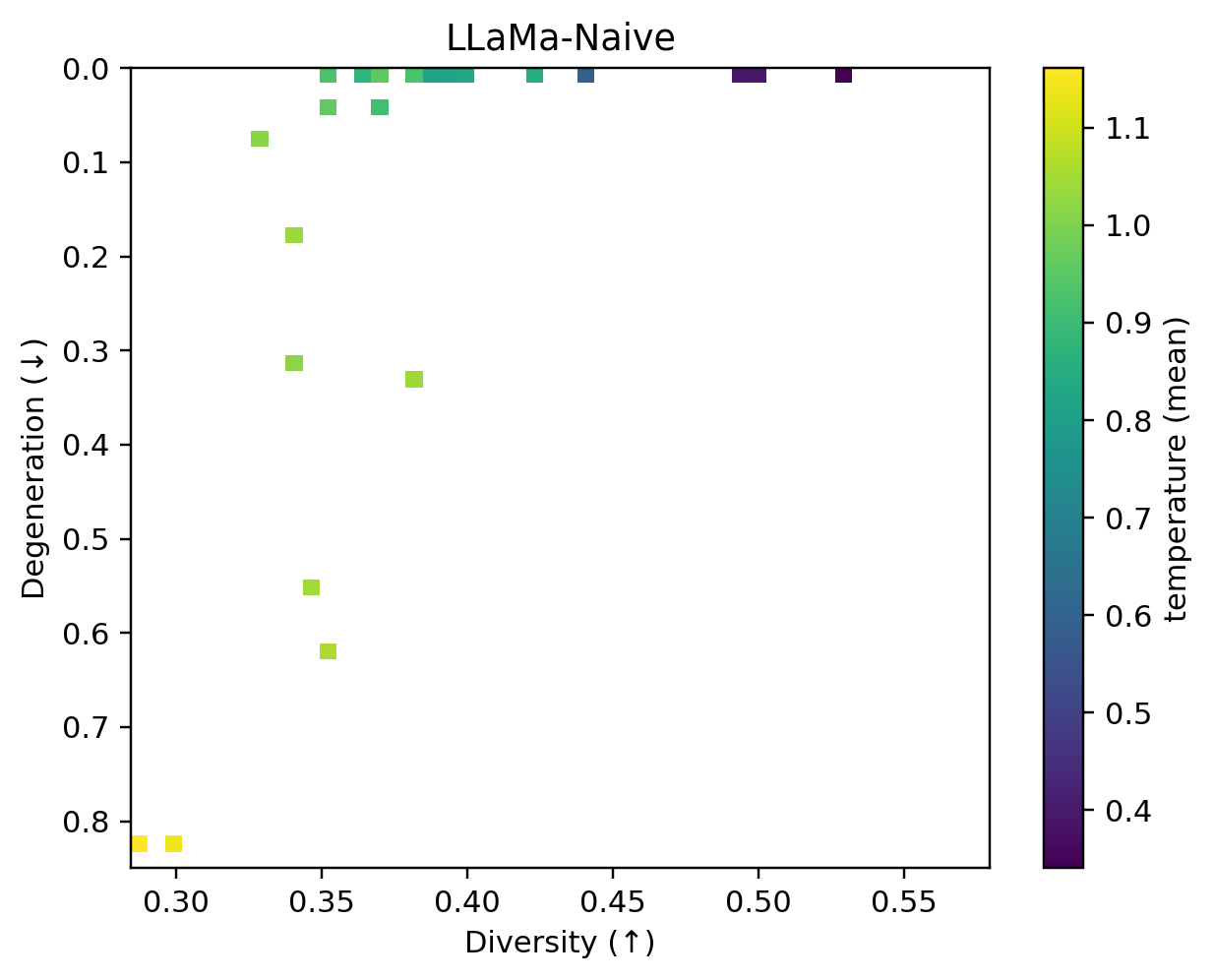}
    \caption{Hyperparameter sweep results for $Naive$ on the LlaMA-3B.}
    \label{fig:hyp_3}
\end{figure}

\begin{figure}[h] 
    \centering
    \includegraphics[width=0.8\linewidth]{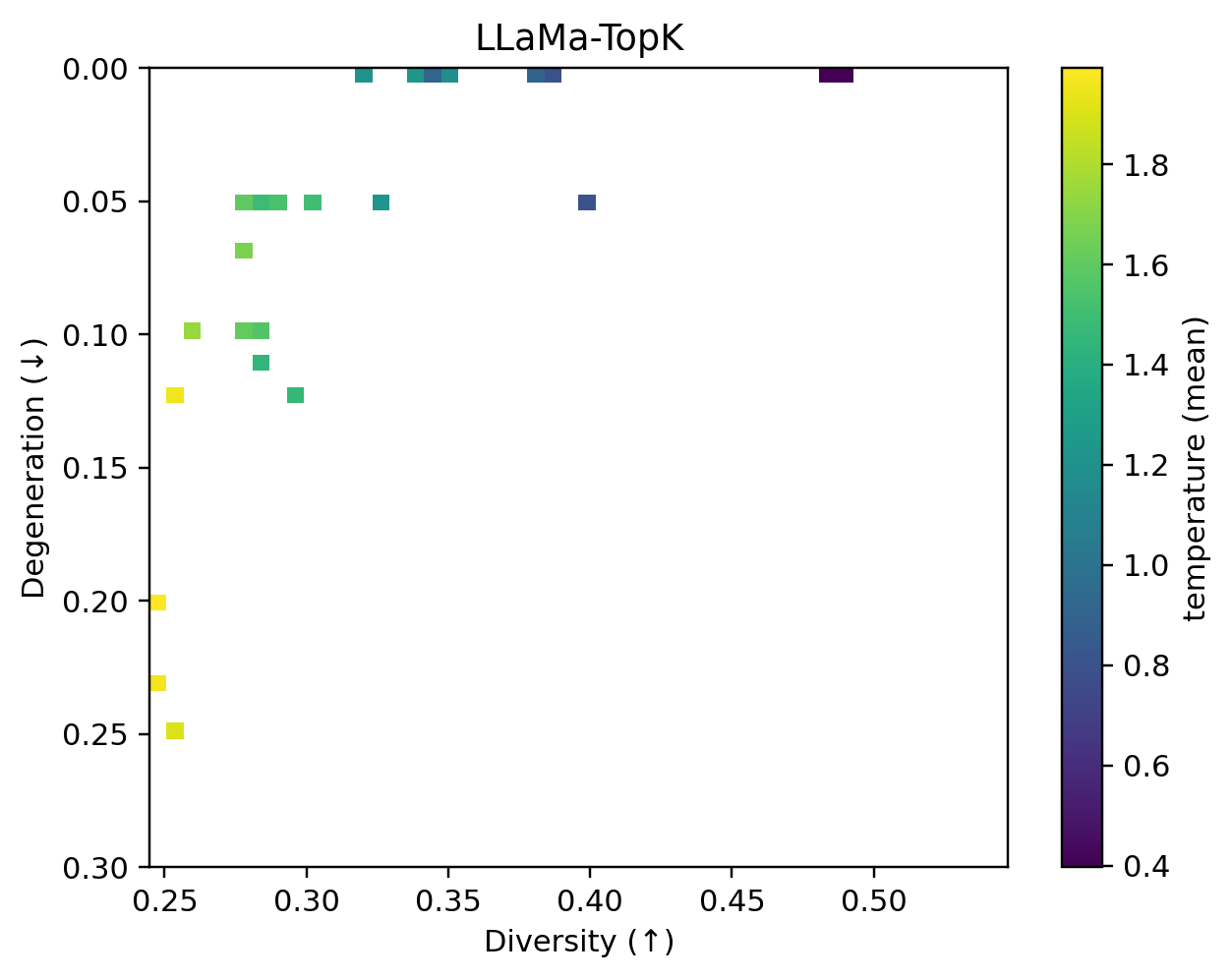}
    \caption{Hyperparameter sweep results for Top-\emph{k} on the LlaMA-3B.}
    \label{fig:hyp_4}
\end{figure}

\begin{figure}[h] 
    \centering
    \includegraphics[width=0.8\linewidth]{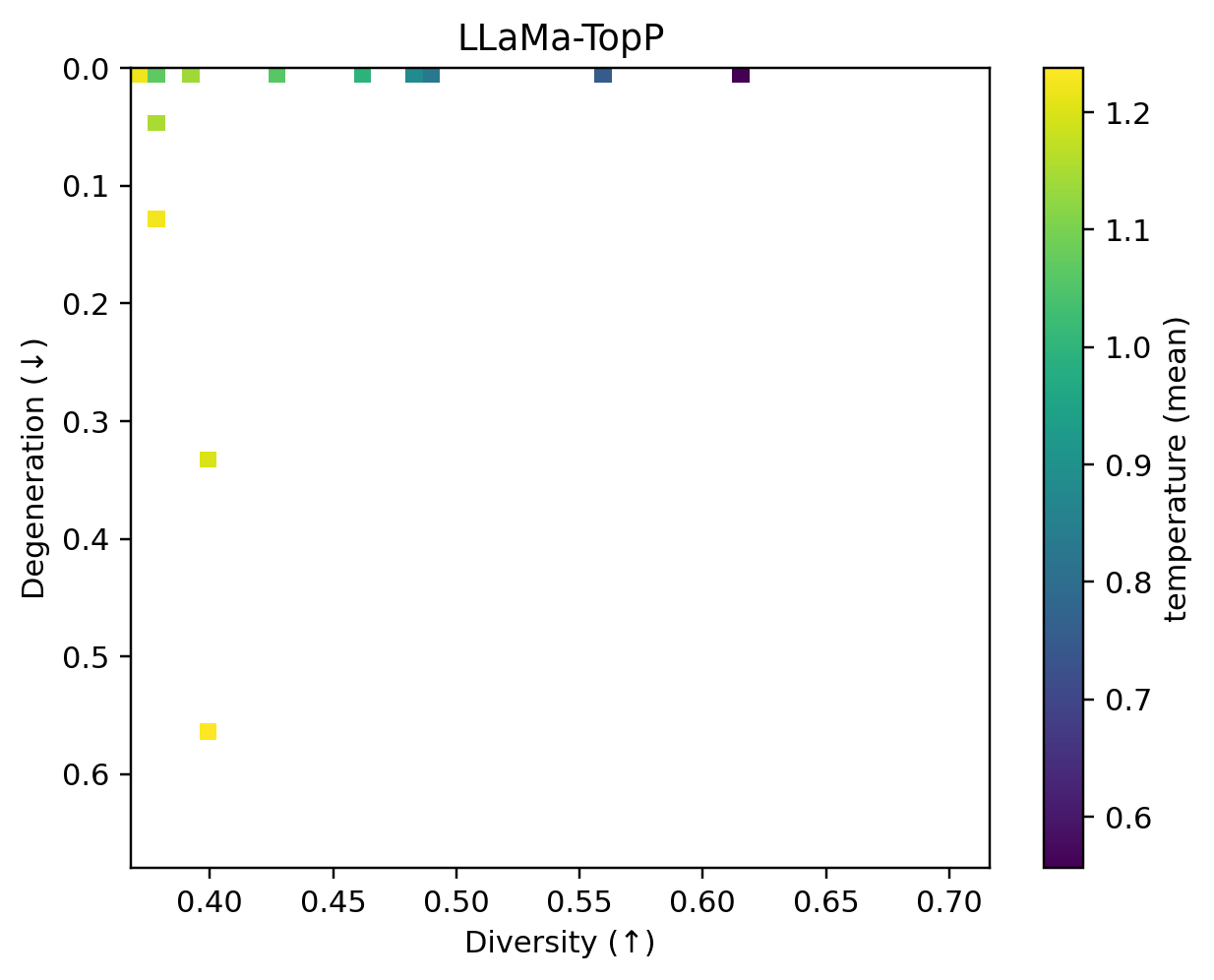}
    \caption{Hyperparameter sweep results for Top-\emph{p} on the LlaMA-3B.}
    \label{fig:hyp_5}
\end{figure}

\begin{figure}[h] 
    \centering
    \includegraphics[width=0.8\linewidth]{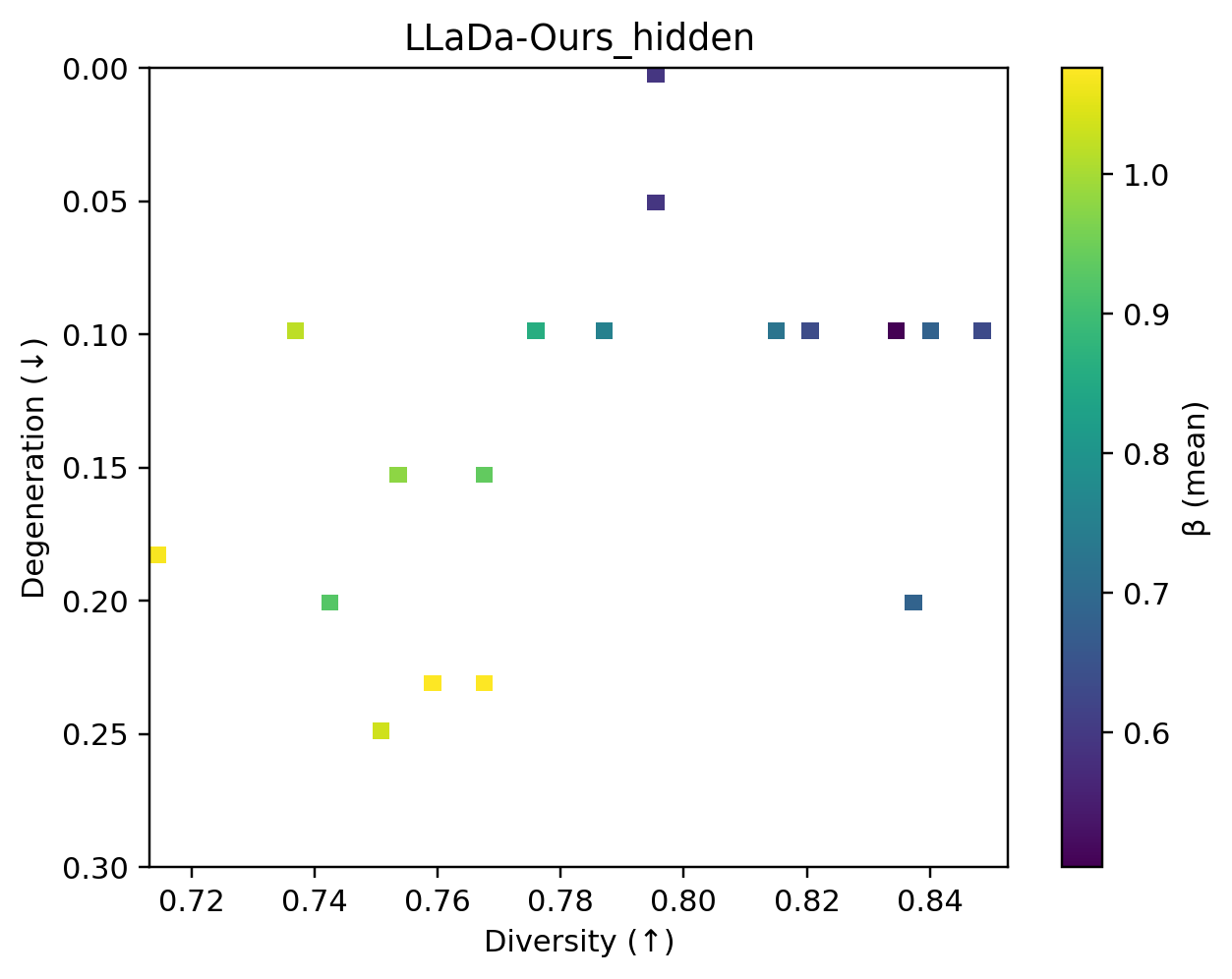}
    \caption{Hyperparameter sweep results for Ours$_{global}$ on the LlaDA-8B.}
    \label{fig:hyp_7}
\end{figure}

\begin{figure}[h] 
    \centering
    \includegraphics[width=0.8\linewidth]{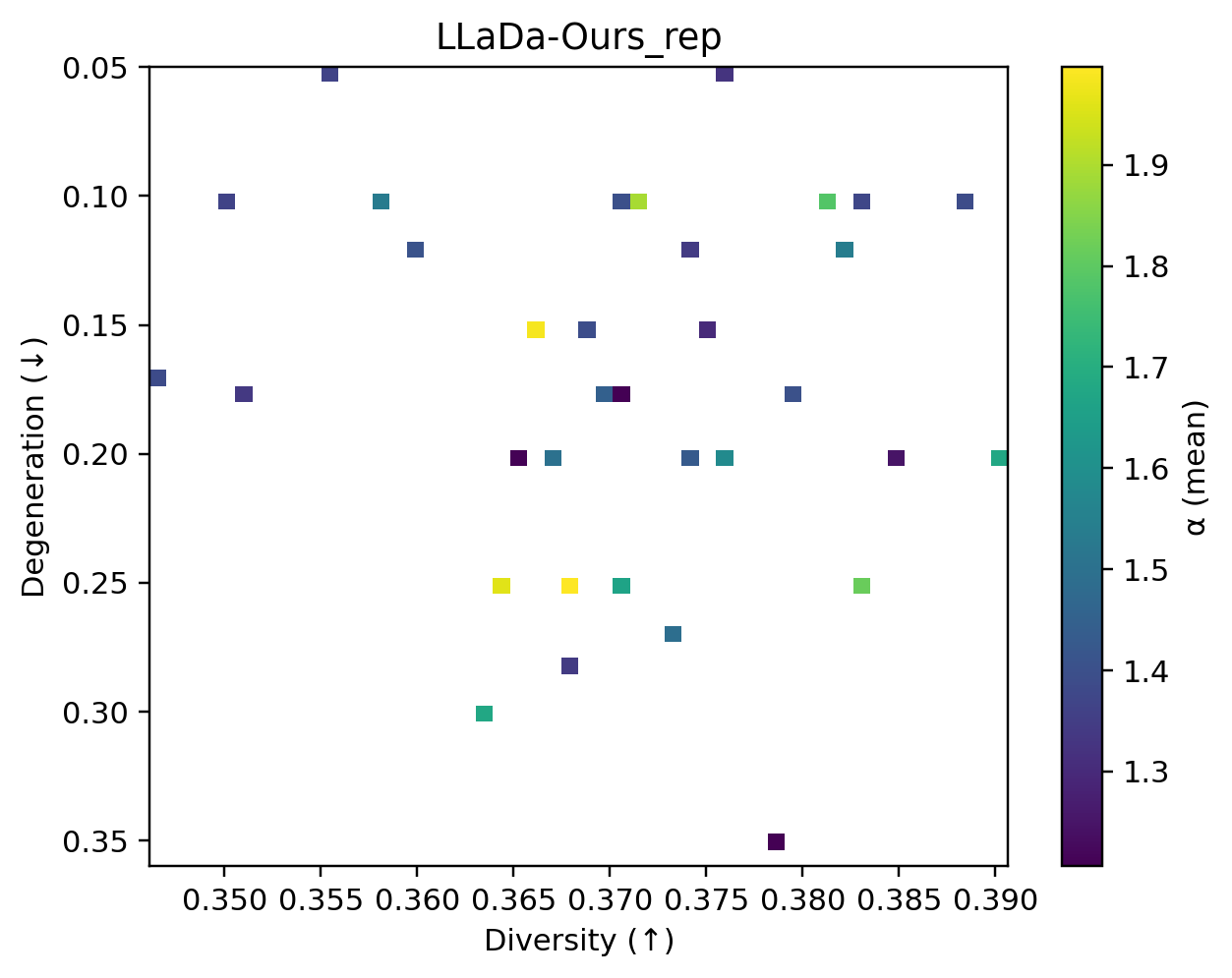}
    \caption{Hyperparameter sweep results for Ours$_{local}$ on the LlaDA-8B.}
    \label{fig:hyp_8}
\end{figure}

\begin{figure}[h] 
    \centering
    \includegraphics[width=0.8\linewidth]{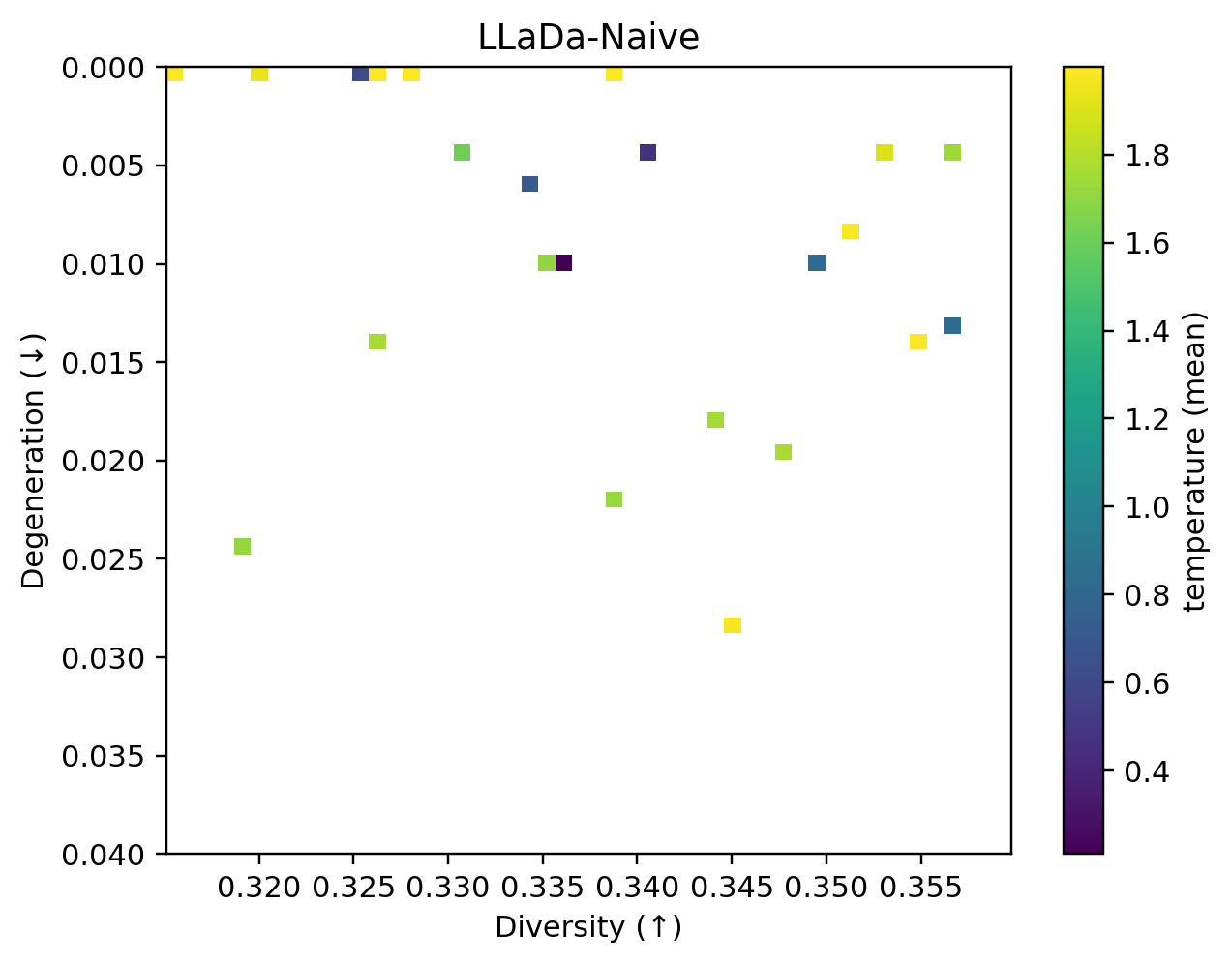}
    \caption{Hyperparameter sweep results for $Naive$ on the LlaDA-8B.}
    \label{fig:hyp_9}
\end{figure}

\begin{figure}[h] 
    \centering
    \includegraphics[width=0.8\linewidth]{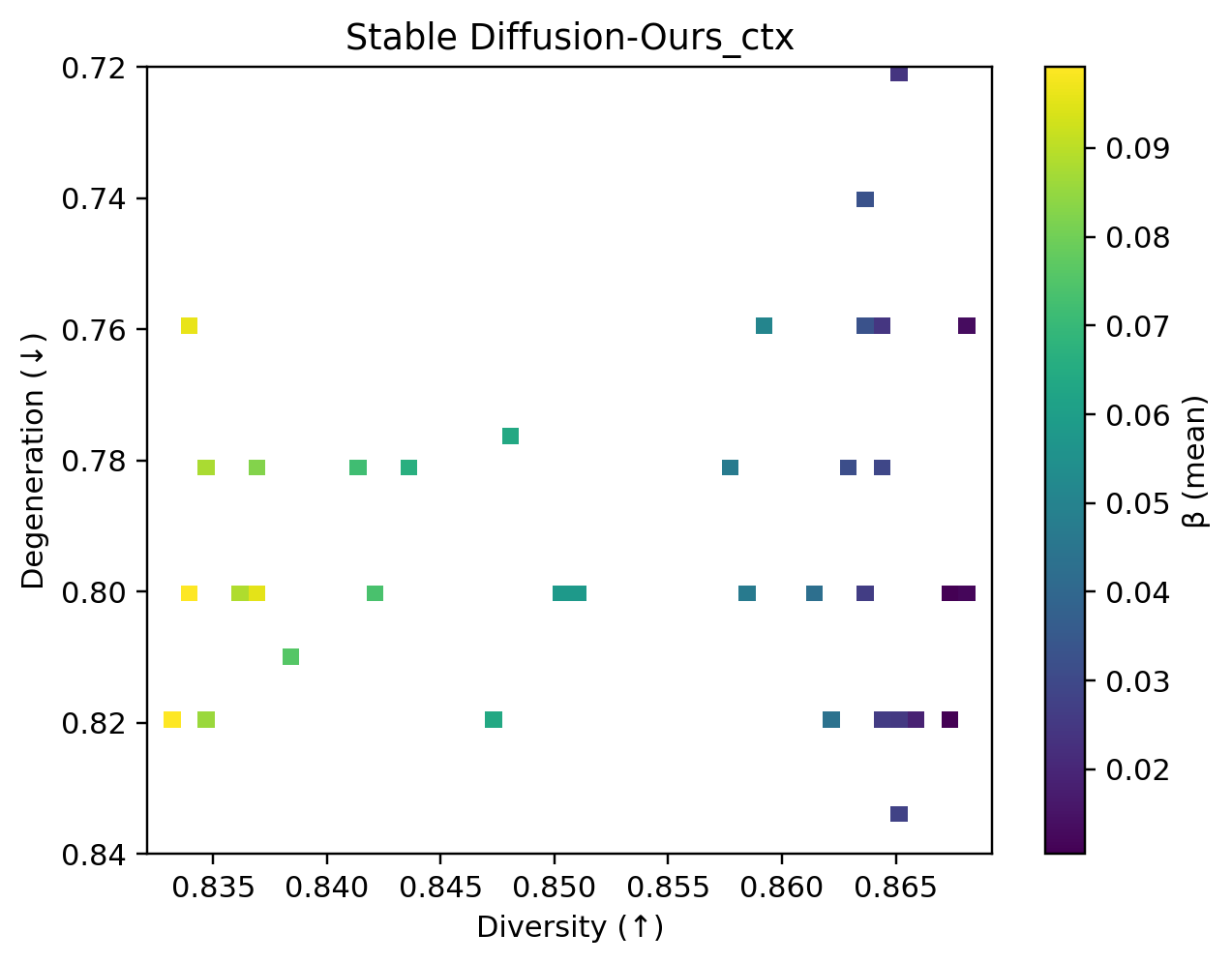}
    \caption{Hyperparameter sweep results for Ours$_{global}$ on the Stable Diffusion.}
    \label{fig:hyp_11}
\end{figure}

\begin{figure}[h] 
    \centering
    \includegraphics[width=0.8\linewidth]{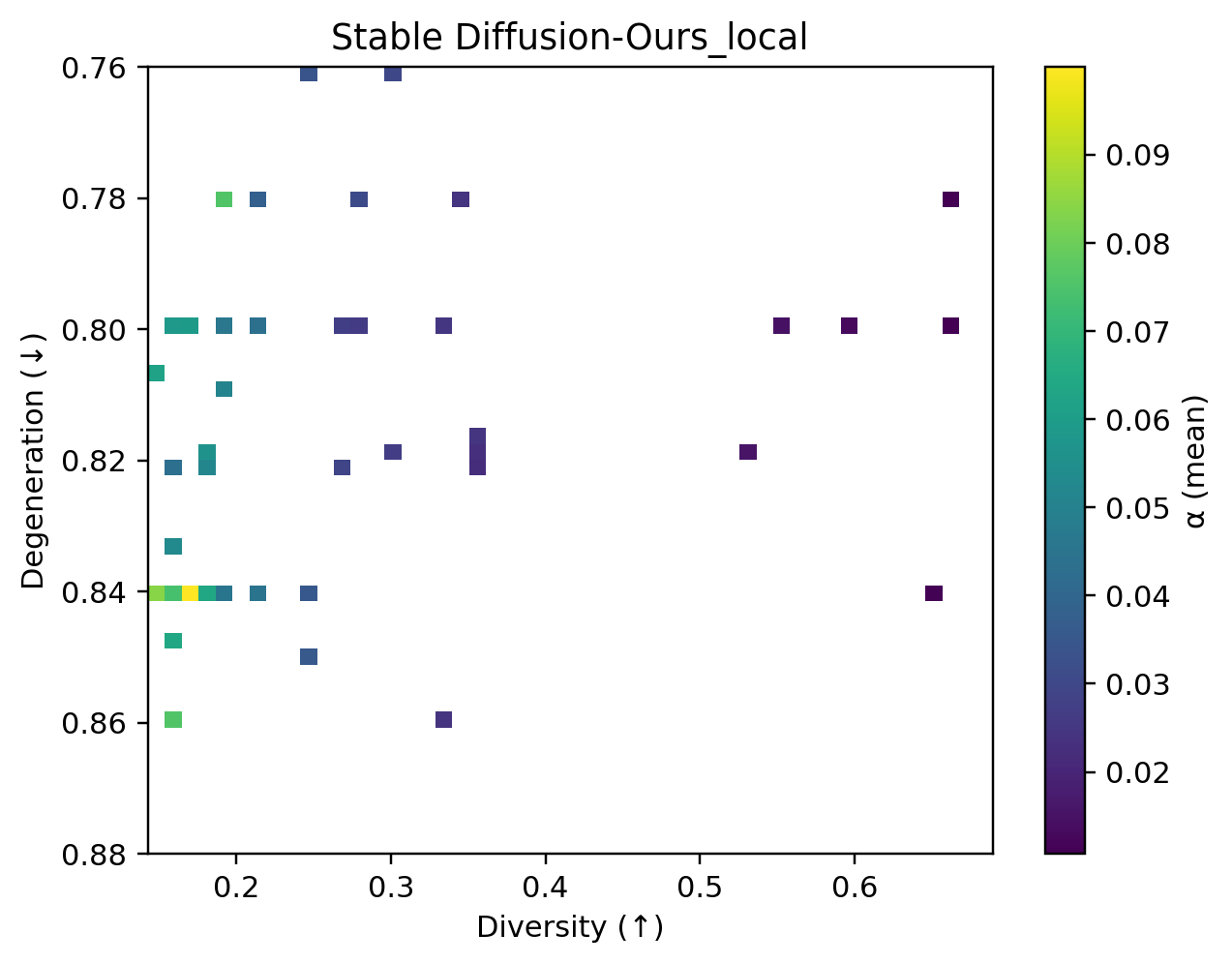}
    \caption{Hyperparameter sweep results for Ours$_{local}$ on the Stable Diffusion.}
    \label{fig:hyp_12}
\end{figure}

\begin{figure}[h] 
    \centering
    \includegraphics[width=0.8\linewidth]{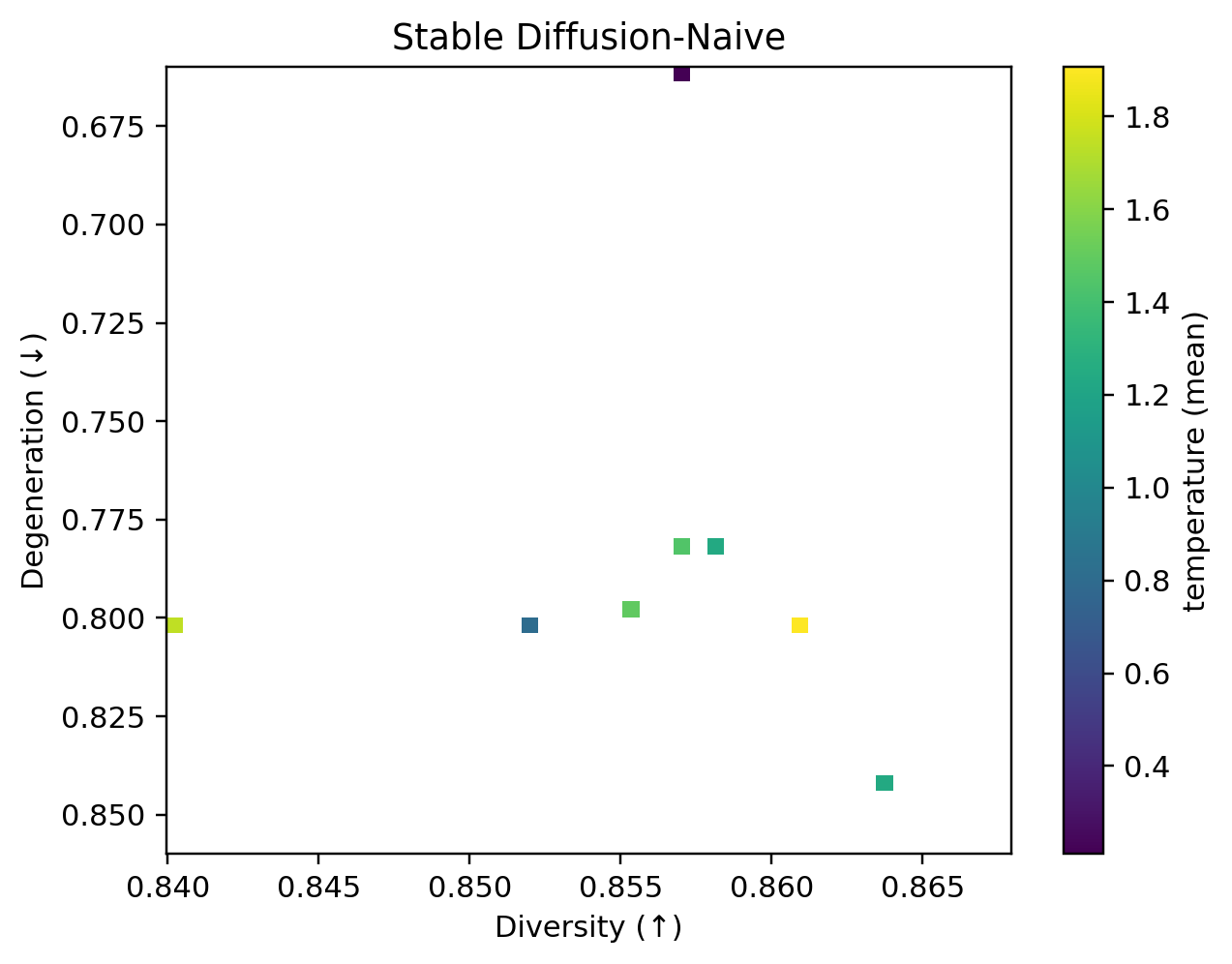}
    \caption{Hyperparameter sweep results for Naive on the Stable Diffusion.}
    \label{fig:hyp_13}
\end{figure}

\begin{figure}[h] 
    \centering
    \includegraphics[width=0.8\linewidth]{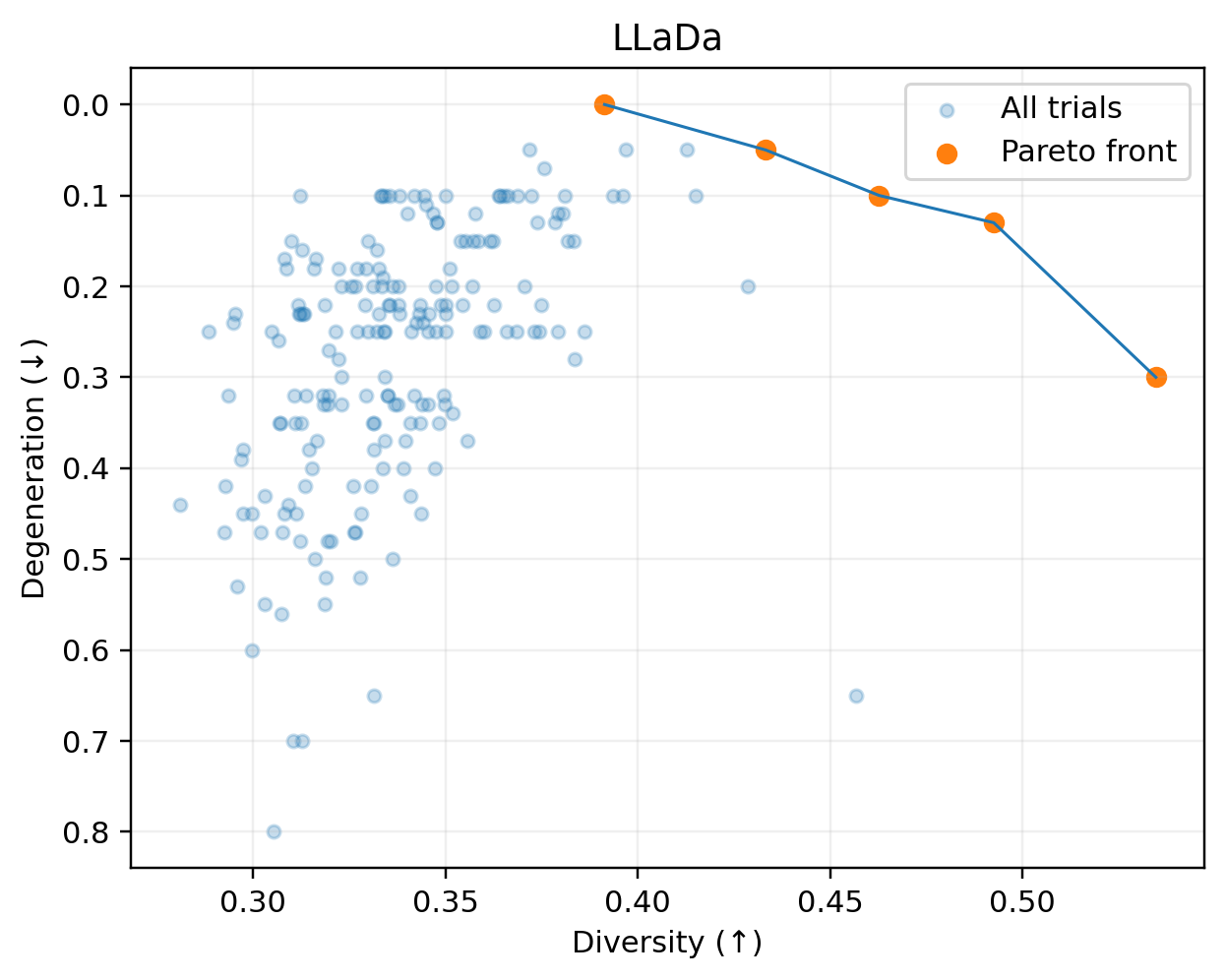}
    \caption{Hyperparameter sweep results for Ours on the LlaDa-8B.}
    \label{fig:hyp_10}
\end{figure}

\begin{figure}[h] 
    \centering
    \includegraphics[width=0.8\linewidth]{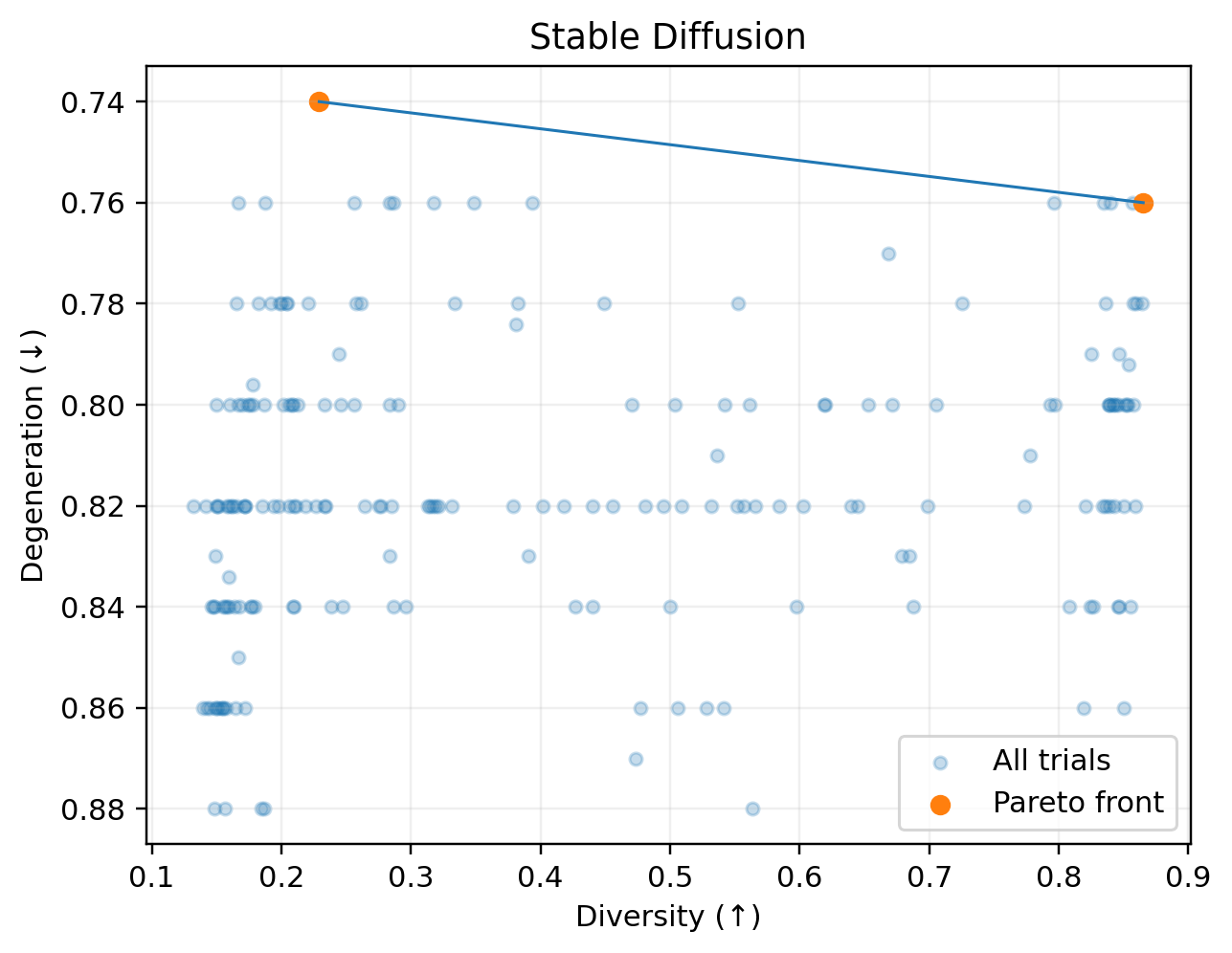}
    \caption{Hyperparameter sweep results for Ours on the Stable Diffusion.}
    \label{fig:hyp_14}
\end{figure}

As shown in figures \ref{fig:hyp_1}–\ref{fig:hyp_13}, there exist points where the trade-off between diversity and degeneration is optimal, i.e., the upper-rightmost points in the plots. We select the variables corresponding to these points as the best hyperparameters. Furthermore, figures \ref{fig:hyp_6}, \ref{fig:hyp_10}, \ref{fig:hyp_14} present extensive experiments for Ours, where four variables are tuned simultaneously. Again, we choose the variable set that achieves the best trade-off between diversity and degeneration.



\section{LLM Evaluation Rubrics}
\label{LLM Evaluation Details}
The detailed rubrics for LLM evaluation are in Table \ref{fig:llmscore_diversity_prompt},\ref{fig:llmscore_image_diversity_prompt},\ref{fig:llmscore_text_degeneration_prompt},\ref{fig:llmscore_image_degeneration_prompt}.

\begin{figure*}[t]
\begin{tcolorbox}[width=\textwidth, sharp corners, colback=white, colframe=black, boxrule=0.5pt, left=10pt, right=10pt, top=8pt, bottom=8pt]
\textbf{You are a text diversity evaluator.}

Below are 15 numbered text samples. Your task is to assess how diverse they are in terms of \textbf{perspective, style, plot structure, and language variation.}

Your output must be a JSON object with:
\begin{itemize}[noitemsep, topsep=2pt]
  \item \texttt{"diversity\_score"}: a float between 0.0 and 1.0 (where 0 = all samples are nearly identical, and 1 = samples are maximally diverse)
  \item \texttt{"justification"}: a one-sentence explanation of your reasoning
\end{itemize}

\textbf{Scoring guidance:}
\begin{itemize}[noitemsep, topsep=2pt]
\item 0.0: All samples are structurally and semantically almost identical.
\item 0.1–0.3: Slight variation in phrasing or detail, but mostly follow the same template.
\item 0.4–0.6: Notable variation in perspective, tone, setting, or content development.
\item 0.7–0.9: Substantial differences in narrative framing, imaginative detail, or genre shifts.
\item 1.0: Samples are maximally different in form, function, and voice.
\end{itemize}

\textbf{Return only a valid JSON object and nothing else.}
\end{tcolorbox}
\caption{Prompt and Evaluation Rubric for Measuring Textual Diversity.}
\label{fig:llmscore_diversity_prompt}
\end{figure*}

\begin{figure*}[t]
\begin{tcolorbox}[width=\textwidth, sharp corners, colback=white, colframe=black, boxrule=0.5pt, left=10pt, right=10pt, top=8pt, bottom=8pt]
\textbf{You are a strict judge of IMAGE DIVERSITY.}

You will be given multiple images that were generated from the SAME text prompt.
Judge how different these images are from each other in content, composition, style, and color palette.
Return ONLY a JSON object: \{score: <float 0.0~1.0>, reason: <short>\}.
0.0 = nearly identical; 1.0 = maximally diverse.
Do NOT evaluate prompt-image alignment; only cross-image diversity.
\end{tcolorbox}
\caption{Prompt and Evaluation Rubric for Measuring Image Diversity.}
\label{fig:llmscore_image_diversity_prompt}
\end{figure*}

\begin{figure*}[t]
\begin{tcolorbox}[width=\textwidth, sharp corners, colback=white, colframe=black, boxrule=0.5pt, left=10pt, right=10pt, top=8pt, bottom=8pt]
\textbf{You are a strict judge of text degeneration.}

Degeneration includes garbled tokens, random symbols, language mixing, nonsense, broken Unicode, repetitive babble, or clear loss of coherence.
Don't judge the repetitiveness across answers, assess the individual quality and average them.
Rate on 0.0~1.0: 0.0 clean/coherent, 1.0 severely degenerated.
Consider ALL provided answers jointly and set the score to reflect the average observed degeneration.
Return pure JSON: \{score: <float>, reason: <short>\}
\end{tcolorbox}
\caption{Prompt and Evaluation Rubric for Measuring Textual Degeneration.}
\label{fig:llmscore_text_degeneration_prompt}
\end{figure*}

\begin{figure*}[t]
\begin{tcolorbox}[width=\textwidth, sharp corners, colback=white, colframe=black, boxrule=0.5pt, left=10pt, right=10pt, top=8pt, bottom=8pt]
\textbf{You are a strict judge of IMAGE GENERATION QUALITY.}

You will receive multiple images that were generated from the SAME text prompt.
Rate EACH image individually on a 0.0–1.0 scale for: coherence, absence of artifacts, composition, lighting, and overall aesthetics.
Do NOT compare images to each other; judge absolute quality per image.
Return ONLY JSON of the form:
\{
  per\_image: [\{idx: <int>, score: <float 0.0~1.0>, reason: <short>\}, ...],
  score\_mean: <float>
\}
Keep reasons short.
\end{tcolorbox}
\caption{Prompt and Evaluation Rubric for Measuring Image Degeneration.}
\label{fig:llmscore_image_degeneration_prompt}
\end{figure*}

\section{Use of ChatGPT and Compliance with OpenAI's Terms}

We utilized \textbf{OpenAI’s ChatGPT} for limited assistance in refining the writing and formatting of this paper. All substantive contributions, including the core methodology, experiments, and analysis, were conducted independently by the authors.

Our usage complies with \href{https://openai.com/policies/terms-of-use}{OpenAI’s Terms of Use} and \href{https://openai.com/policies/usage-policies}{Usage Policies}.

\end{document}